\documentclass[journal,dvipsnames]{IEEEtran}

\usepackage{epsf}
\usepackage{graphicx}
\usepackage{hyperref}
\usepackage{xr-hyper}
\usepackage[ruled,vlined]{algorithm2e}
\usepackage{booktabs, makecell, tabularx}
\usepackage{amssymb}
\usepackage{amsmath}
\usepackage{mathtools}
\usepackage[table]{xcolor}    
\usepackage{comment}
\usepackage[labelsep=period]{caption}
\usepackage{soul} 
\renewcommand{\thetable}{\arabic{table}}

\begin{document}
\title{Parameterless Gene-pool Optimal Mixing Evolutionary Algorithms}
\title{Parameterless Gene-pool Optimal Mixing Evolutionary Algorithms}
\author{Arkadiy Dushatskiy, Marco Virgolin, Anton Bouter, Dirk Thierens and Peter~A.~N.~Bosman%
\thanks{Arkadiy Dushatskiy is with the Life Sciences and Health research group of the Centrum Wiskunde and Informatica}
\thanks{Marco Virgolin is with the Life Sciences and Health research group of the Centrum Wiskunde and Informatica}
\thanks{Anton Bouter is with the Life Sciences and Health research group of the Centrum Wiskunde and Informatica}
\thanks{D. Thierens is with the department of Information and Computing of the University of Utrecht}
\thanks{Peter~A.~N.~Bosman is with the Life Sciences and Health research group of the Centrum Wiskunde and Informatica and the department of Software Technology of the Delft University of Technology}
}
\maketitle

\begin{abstract}
When it comes to solving optimization problems with evolutionary algorithms (EAs) in a reliable and scalable manner, detecting and exploiting linkage
information, i.e.,~dependencies between variables, can be key. \\
In this article, we present the latest version of, and propose substantial enhancements to, the Gene-pool Optimal Mixing Evoutionary Algorithm (GOMEA): an EA explicitly designed to estimate and exploit linkage information. 
We begin by performing a large-scale search over several GOMEA design choices, to understand what matters most and obtain a generally best-performing version of the algorithm. Next, we introduce a novel version of GOMEA, called CGOMEA, where linkage-based variation is further improved by filtering solution mating based on conditional dependencies. \\
We compare our latest version of GOMEA, the newly introduced CGOMEA, and another contending linkage-aware EA DSMGA-II in an extensive experimental evaluation, involving a benchmark set of 9 black-box problems that can only be solved efficiently if their inherent dependency structure is unveiled and exploited.
Finally, in an attempt to make EAs more usable and resilient to parameter choices, we investigate the performance of different automatic population management schemes for GOMEA and CGOMEA, \emph{de facto} making the EAs parameterless. 
Our results show that GOMEA and CGOMEA significantly outperform the original GOMEA and DSMGA-II on most problems, setting a new state of the art for the field.
\end{abstract}


\IEEEpeerreviewmaketitle

\section{Introduction}
\IEEEPARstart{K}{ey} to the success of any optimization algorithm in terms of
search effectiveness and efficiency, is the ability to exploit structural
features of the problem being solved. 
To this, Evolutionary Algorithms (EAs) are no exception. 
For EAs, it is predominantly the variation operators that need to be favorably configured to exploit structural features. 
One structural feature that is of particular importance is variable dependence. 
Not only does variable dependence have a direct influence on the inherent difficulty of a problem, not being able to exploit such dependency information may lead to very inefficient optimization performance. 
If two variables are completely independent in a problem, this problem can be solved by considering the variables separately. 
Conversely, if two variables are strongly
dependent, joint settings of these variables need to be considered in order to
find the optimal solution. In EAs, such dependencies (between the variables that
are directly manipulated by the EA, i.e., the genes), are also known as
\emph{linkages}. It has been long known that groups of variables that exhibit
such strong linkages need to be treated, with high probability, in a joint fashion by the variation operator in order for an EA to be an efficient solver \cite{thierens1999scalability, pelikan1999boa}.
Especially in the domain of discrete variables that constitute a Cartesian
search space, which is also the domain that this article pertains to, many EAs
employ a mixing operator that exchanges parts of solutions. Ensuring this mixing
operator is linkage-friendly, i.e.,~has a high probability of exchanging groups
of genes that are highly dependent, can make the difference between obtaining
efficient, i.e.,~low-polynomial, and inefficient, i.e.,~exponential, scale-up of
the required running time to solve the problem~\cite{thierens1999scalability, pelikan1999boa}.

The relevance and importance of linkage processing is even more prominent when
taking a black-box perspective on optimization. In Black-Box Optimization (BBO),
there is very little to no information available on the problem being solved.
Metaheuristics, among which EAs, are commonly formulated and studied in this
context, with the notion of designing a powerful general problem solver in mind.
Certainly, the no-free-lunch theorem assures us, that considering all possible
optimization problems, no such solver exists~\cite{wolpert1997no}. However, a
generally valid assumption can be made that the types of optimization problems
we are interested in, are not completely random, but have some sort of
exploitable structure. It is the exploitation of this structure however that
governs whether or not optimization will proceed effectively and efficiently.
This then brings us back to the linkage problem, for it is assumed that the
structure of the typical optimization problems we are interested in, is
nontrivial, i.e.,~its variables are not all fully independent. For this reason,
we have no guarantee that a simple genetic algorithm with uniform crossover, or
any static crossover operator for that matter, will effectively exploit the
structure of the problem. Thus, their use comes with the risk of exponential
scale-up of the required runtime on problems that are polynomial-time
solvable~\cite{thierens1999scalability}. To avoid this, linkage information needs to be exploited
properly. In a BBO setting however, such information isn't readily available,
and thus must be determined otherwise, using previously performed
solution-quality, i.e.,~fitness, evaluations. This process is commonly known as
\emph{linkage learning}, which is a key concept in this article.

An argument can be made at this point that the added complexity and effort of
performing linkage learning is superfluous because a true BBO scenario isn't frequently encountered when solving real-world problems. 
A need for BBO may still very well surface however, even when efficient local search
heuristics are available for a particular problem. It is well-known that
combining EAs and local search is highly effective for many
problems~\cite{hart2005memetic}.
The reason for this is that by applying local search to every solution, a second search problem can be seen to exist in the space of local
optima of the optimization problem. Running a local search heuristic multiple
times, i.e.,~a random restart heuristic, then can effectively be seen as random
search in the space of local optima. This space may be searched more efficiently
using an EA, which can be obtained by applying local search
to every solution that the EA generates. Even when we understand very well the
problem being solved to the point where we can design efficient local search
algorithms, the nature of the search space composed of the local optima may
still be extremely hard to analyze. In that space then, there is again a need
for powerful BBO algorithms.

The linkage problem has already been identified a long time ago, and much work
on tackling this problem has previously been done, often presented
simultaneously with a new EA, see e.g., \cite{pelikan2006hierarchical, bosman2012linkage}. Much of this work has been toward building more
complex models that are capable of capturing problem structure in more intricate
detail, up to the relatively complex task of estimating entire (factorized)
probability distributions, as is done in Estimation-of-Distribution Algorithms
(EDAs)~\cite{larranaga2001estimation}. Although ultimately capable of exploiting problem structure
properly, the overhead involved with estimating actual probabilities surpasses
the need to determine the linkage information that needs to be effectively
exploited. Importantly, such overhead becomes more significant on large-scale problems causing scalability issues. This article focuses specifically on the linkage hurdle on the road
to powerful BBO algorithms. In particular, we introduce the new version of the Genepool Optimal Mixing Evolutionary Algorithm (GOMEA) that seamlessly integrates the
traditionally separate operators of selection and variation in EAs in order to
get the most out of available linkage information. Moreover, a generalized model
of linkage information allows linkage information to be processed at more than
one level, e.g.~processing a hierarchy of weak and strong dependencies. 

This article extends our previously published work on GOMEA by a more extensive experimental analysis on more optimization problems and larger problem sizes, testing the impact of various possible design choices such as local search operators on GOMEA, and, importantly, we propose a novel variation operator which exploits conditional dependencies between sets of variables and is called Conditional GOM (CGOM). 
Finally, we demonstrate the practical applicability of GOMEA by designing parameterless modifications of it. The performance of new and old versions of GOMEA are compared is shown in comparison with other EAs, including, the
recent version of DSMGA-II \cite{chen2017two}, and the Parameterless Population Pyramid (P3) \cite{goldman2015fast}. 
Moreover, this article joins all algorithmic information from our previous work that is needed to make this article self-contained and represent the current state-of-the-art in the GOMEA research line.

The remainder of this article is organized as follows. In
Section~\ref{sectionrelatedwork} we discuss related work. In
Section~\ref{sec:sectiongomeafamily} we outline the general working scheme of GOMEA and present design options for its most important
components in more detail. Also we present GOMEA instances without the population size parameter and describe schemes to run GOMEA in a population size-free fashion. 
Then, we present our benchmark problems and the design of experiments in
Section~\ref{sec:sectionexperiments}, followed by the results in Section~\ref{sec:Results}. The article ends with a discussion in Section~\ref{sec:discussion} and conclusions in Section~\ref{sec:conclusion}.

\section{From Genetic Algorithms\\ to Estimation-of-Distribution Algorithms\\ and Back Again}\label{sectionrelatedwork}
It was already hypothesized by John Holland himself that the simple Genetic
Algorithm succeeds at optimization if it can proliferate important building blocks \cite{holland1992adaptation},
i.e.,~partially defined solutions for which it holds that, when averaging over all the solutions that it is part of in a population, the fitness is better than the average fitness of the population.

Key is the proper mixing of these partial
solutions, which means disrupting them as little as possible (i.e.,~copying them entirely from one solution to the next) and not copying other parts of the
solution that they are (semi)-independent from. 
If these important partial solutions have a large probability of being destroyed during variation, for instance by using uniform crossover,
the population size that is required to find the optimum may grow exponentially with the problem size. 
Conversely, polynomial population size growth can be achieved if the partial solutions are properly mixed.
A well-known example of this is represented by the sum of additively decomposable, non-overlapping, deceptive trap functions ~\cite{deb1993analyzing}.

Since then, there has been a dedicated research line in the field of
evolutionary computation to design variation operators that are capable of
automatically detecting the presence of important building blocks, and of
reconfiguring the way in which variation proceeds to ensure that building blocks
are mixed well and disrupted as little as possible. 
The first family of
EAs along this line was the messy genetic algorithm
family~\cite{kargupta1996gene}. 
Algorithms in this family allowed genes to be re-ordered by
explicitly encoding their location. Although eventual algorithms were able to
avoid exponential scale-up, the overhead of re-ordering genes was still
substantial and lacked explanatory statistical underpinning.

For this reason, researchers started looking into probabilistic approaches that were capable of explicitly computing dependencies between problem variables by estimating probability distributions over them. The population can be seen as a database that represents the type of solutions that are desired, and, over time, through selection, gets pushed toward the optimum. 
Selecting the better solutions makes dependencies stand out, since on average the solutions that contain important building blocks will have a better fitness than those solutions that do not. By estimating a probability distribution from the population, these dependencies can be explicitly modeled in a probabilistic fashion.
Moreover, by sampling new solutions from the estimated distribution, these dependencies are respected. Because the process of sampling generates a new database that has the same statistical properties as the original database (to the extent to which these properties were modeled in the probability distribution), this approach can be considered as mixing solutions at a population level rather than at the two-parent level as was before typically reminiscent of genetic algorithms. 
This type of algorithm is known as the
Estimation-of-Distribution Algorithm (EDA)~\cite{larranaga2001estimation,lozano2006towards}. 
Effectively and efficiently estimating probability distributions that capture higher-order dependencies was still key however to avoid exponential scale-up on problems with non-trivial dependency structure. 
Initial attempts that used either univariately factorized
probability distributions (e.g.,~pBIL~\cite{baluja1995removing}, and cGA~\cite{harik1999compact}) that modeled every variable to be independent from every other variable, or bivariately factorized distributions that considered variable dependencies of at most order two (e.g.,~MIMIC~\cite{de1997mimic}, COMIT~\cite{baluja1997combining}, and BMDA~\cite{pelikan1999bivariate}) still fail to obtain polynomial scale-up on the additively decomposable deceptive trap functions.
Low-order polynomial scale-up is only obtained by EDAs that model higher-order dependencies (e.g.,ECGA~\cite{harik2006linkage}, LFDA~\cite{muhlenbein1999fda}, and BOA~\cite{pelikan1999boa}).

The most advanced EDA in this line is commonly accepted to be the hierarchical Bayesian Optimization Algorithm (hBOA)~\cite{pelikan2006hierarchical}.
Both its predecessor BOA and hBOA itself estimate a Bayesian network every generation using a greedy learning procedure. 
hBOA however has the ability to store the parameters in this network more efficiently by storing only those combinations of values for dependent variables that actually appear in the population, thereby preventing the need to generate huge probability tables that require the explicit enumeration of \emph{all} possible value combinations of a
set of dependent variables. 
This, combined with a mechanism (restricted tournament selection) that promotes population diversity, allowed hBOA to be the only EDA capable of solving problems with hierarchical dependency structures while requiring only low-order polynomial-time scale-up of the population size and number of function evaluations 
Although to a large extent now satisfactorily
solving the linkage problem and providing a solid, statistically sound basis for
doing so, the overhead required by hBOA is still substantial, requiring
asymptotically $\mathcal{O}(n\ell^{3})$ time per generation where $\ell$ is the number
of problem variables and $n$ the population size. Moreover, the number of
generations required to solve a problem is typically in the same order as a
properly configured GA, which is typically in the order of
$\mathcal{O}(\sqrt{\ell})$ \cite{pelikan2005hierarchical}.

Although a solid approach to tackling the linkage problem, estimating entire probability distributions comes with the necessity to estimate not only a dependency structure, but also to estimate parameters (e.g.,~actual probabilities). Moreover, in order to decide what underlying dependency structure is a good one, i.e.,~not missing key dependencies and not overly complex, quality-of-fit measures need to be computed that decide when to stop the greedy learning approach that iteratively increases the complexity of the underlying dependency structure. 
These aspects are not necessarily important for
tackling the linkage problem, because for that it would suffice to know which variables are (strongly) dependent on which other variables. 
The joint probabilities of entire building blocks do not explicitly need to be computed, as they are stored implicitly in the population.
Mixing the information stored in the population therefore automatically follows these probabilities. 
It is on these foundations that the GOMEA framework is based. 
GOMEA was first introduced in 2011~\cite{bosman2012linkage}, posed as a broadened scope of the idea behind the original
Linkage Tree Genetic Algorithm (LTGA) introduced in 2010~\cite{thierens2011optimal}. 
LTGA was one of the first algorithms to depart from the EDA principle of estimating entire probability distributions, and thus essentially going back to the notion of genetic algorithm, but still using similar statistical concepts as used in EDAs to detect dependencies. 
Ultimately this lead to a model-building complexity of an order of magnitude faster ($\mathcal{O}(n\ell^{2})$) than hBOA, while being able to
capture and exploit both low-order dependencies as well as high-order
dependencies at the same time. 
Moreover, LTGA requires only a handful of generations to find the optimal solution due to much more extensive model exploitation during variation, further reducing the overall required model-building complexity. 
As later versions of LTGA, including the one presented in this article, are seen as instances of the GOMEA framework, details will be described in subsequent sections. 
Besides LTGA which we will from now refer to as LT-GOMEA, other non-EDA algorithms that build models to model and exploit linkage information have been proposed of late~\cite{dsmga2, chen2017two, goldman2015fast}.  
These algorithms that can more generally be described as model-based EAs have also been successful at outperforming hBOA.

\section{The GOMEA Family of EAs} \label{sec:sectiongomeafamily}

The family of Gene-pool Optimal Mixing Evolutionary Algorithms (GOMEAs) has been proved to show impressive performance on benchmarks and, importantly, real-world problems. For instance, Real-Valued Multi-Objective GOMEA (RV-MO-GOMEA) \cite{bouter2019gpu} is now used for brachytherapy treatment planning optimization. This application received a Silver Humies award \cite{humies} which highlights its practical value and outstanding, better-than-human performance. Another example is adaptation of GOMEA for Genetic Programming (GP-GOMEA) \cite{virgolin2017scalable}. Beside showing better performance than alternative GP algorithms on classical machine learning benchmarks, GP-GOMEA has been also successfully applied to a real-world medical problem, namely, a radiotherapy dose reconstruction \cite{virgolin2018symbolic, virgolin2020machine}. This application was noted with a Silver Humies award in 2021. These two examples show the potential of the GOMEA family of algorithms.

The family of GOMEAs is actually a subset of the OMEA family~\cite{bosman2012linkage}. Another subset is the Recombinative OMEA (ROMEA) family whereby mixing of solutions occurs only between 2 parent solutions rather than between all solutions in the population as is the case for GOMEA. 
When tested on various problems however, GOMEA was found to have the best performance as long as the models capturing linkage information were adequate \cite{bosman2012linkage}. For this reason, we focus particularly on GOMEA here.

The main idea behind the OMEA framework is that linkages are identified using sets of variable indices (see Section~\ref{subsectionfos}), which we shall also call linkage sets. 
These individual linkage sets are then explicitly exploited rather than implicitly as is the case in classical GAs and EDAs. In the latter,
entire solutions are generated and subsequently evaluated. The main idea of OMEA however is to take values only for a linkage subset from a donor solution, and try these values out in another solution to see if it improves. 
It is this direct notion of acceptance that makes the success of the mixing operation independent of the effect of all other mixing events that may happen when constructing an entire new solution first. 
Because this makes each mixing event an optimal decision, unhampered by potential collateral noise, and because when all linkage sets are correctly identified, mixing essentially does not make any mistakes this way (unless unhelpful donor solutions are selected), this approach to variation was called Optimal Mixing (OM).

GOMEAs are a subclass of the general class of EAs and as such are a form of population-based search. 
The most traditional approach to population management is to have a population of a fixed size. 
We will discuss what GOMEA looks like with this approach as well as with different approaches to population management that no longer require the specification of a value for the population size parameter. The latter is especially of high practical value. 

In the remainder of this section we provide more details on the various
components of GOMEA.

\subsection{Family Of Subsets (FOS) as a Linkage Model}\label{subsectionfos}
The GOMEA class of EAs focuses on modeling linkage by explicitly identifying
sets of variables to be treated jointly in the variation process. Moreover, such
linkage sets are allowed to overlap. Specifically, \emph{any} subset of the set
of all variables may be identified within the linkage model. This may be defined
as follows. Let $\mathcal{L} = \{0,1,\ldots,\ell-1\}$ be the set of $\ell$ unique
identifiers of variables that the EA processes, then the linkage model in GOMEA
is a subset of the powerset of $\mathcal{L}$. Such a set is commonly called a
family-of-subsets in mathematics. We therefore call the linkage model in GOMEA
the family-of-subsets, or FOS, model, and denote it by $\mathcal{F}$, i.e:

\begin{equation}
  \mathcal{F}
  \subseteq
  \wp(\mathcal{L})
\end{equation}

\subsubsection{Linkage Tree (LT) Model}
Though different ways to configure a FOS model by learning linkage from the
population were introduced, we focus here on a so-called Linkage Tree (LT) model which demonstrated efficiency in solving various combinatorial optimization problems \cite{thierens2011optimal}. An LT is a binary tree with $2\ell-1$ vertices. LT leaves are singletons of problem variables, the root of a LT is the set of all problem variables $L$, and all other vertices are variables subsets $F^i$ which are unions of disjoint subsets of children $k,j$ of vertex $i$: $F^i=F^j \cup F^k, F^j \cap F^k = \emptyset$.


\subsubsection{Similarity Measures}
An LT can be built in a bottom-up fashion using hierarchical clustering \cite{kraskov2009mic}: starting from singletons, the most similar subsets of variables are merged until a subset containing all variables is obtained (a tree root). A similarity between two subsets of variables $F^i$, $F^j$ is defined as average similarity measure of all pairs of variables $(X,Y)$ where $X \in F^i, Y \in F^j$. Different similarity measures can be used \cite{bosman2012measures}. Here, we consider two of them which are most commonly used (e.g., in \cite{luong2018multi, den2016multiple, goldman2015fast}), namely, standard Mutual Information (MI) and Normalized Mutual Information (NMI). For two variables $X,Y$ MI and NMI are defined as 

$$
\text{MI}(X,Y) = \text{H}(X) + \text{H}(Y) - \text{H}(X \cup Y)
$$
$$
\text{NMI}(X,Y) = \text{MI}(X,Y)/\text{H}(X \cup Y )
$$

where $H(X)$ is information entropy, defined as 
$$
H(X) = \sum_{x \in \Omega_X}{-P(X=x)\log(P(X=x))}
$$

\subsubsection{Linkage Tree Filtering}
It was shown in \cite{bosman2013more} that a full LT model (with $2\ell-1$ vertices) may have redundant subsets which can be filtered out to increase mixing efficiency. 
Here we consider one particular case of filtering which was successfully applied in \cite{goldman2015fast}. 
When two subsets $F^j$ and $F^k$ are merged into a subset $F^i$, it may happen, that the similarity between them is maximal (one in case of MI or NMI), which means that in a population, values of variables from one subset can perfectly predict values of variables from another subset. 
We suppose that there is no merit in using these subsets in mixing separately as it may disrupt this pattern and use additional unnecessary evaluations. 
Thus, keeping subsets $F^j$ and $F^k$ in a FOS is not reasonable and it is sufficient to keep only the parent subset $F^i$. 
In practice, to deal with possible numerical errors in similarity measure calculation, the filtering rule is invoked if the similarity measure value is above $1-\varepsilon$ threshold (we use $\varepsilon=10^{-6}$). Let $S(X, Y)$ be the similarity measure.
After the filtering rule is applied, the subsets of an LT model satisfy the description: 

\begin{equation}
\begin{gathered}
    \forall F^i, F^j, F^k \in \mathcal{F} ~\textnormal{s.t.}~ F^i = F^j \cup F^k,  \\
    S(F^j, F^k) \le 1-\varepsilon 
\end{gathered}
\end{equation}

\subsection{Gene-pool Optimal Mixing}
Variation in GOMEA is guided by the contents of the FOS model in order to prevent disrupting the linkage information it represents. 
To do so, an operator called \emph{Gene-pool Optimal Mixing (GOM)} is used that integrates selection and variation and has many similarities with greedy search algorithms. The GOM operator is described in pseudo-code in Algorithm~\ref{alg:GOM}.

GOM is applied to a single solution and outputs a single solution that is never worse than the input solution. To improve a solution, GOM loops over the contents of the FOS model. 
We consider two ways of iterating over FOS elements: in random order \cite{thierens2011optimal} and ascending order of subsets size ($|F^i|$) \cite{goldman2015fast}. 
For each linkage subset $F^i$, GOM attempts to overwrite the values of the variables in $F^i$ of the solution in consideration, with values from a donor solution that is chosen at random from the population.
If this overwriting action does not cause the fitness of the solution to become
worse, the copy action is accepted. Otherwise, the donor material is rejected
and the action is undone. 
To allow traversing of fitness plateaus, changes that lead to the same fitness are also accepted if the solution is not the elitist one.
Note that a FOS subset containing all variables, i.e., the root of the LT model, is not used in GOM, as it implies replacing an entire solution rather than changing only a part of it.
\subsubsection{Exhaustive Donor Search}
When population diversity becomes low, it is likely that a randomly selected donor has the same genes $F^i$ as the current solution, therefore, no new genotype is obtained. 
To deal with this situation, we can continue trying different donors until one is found in which genes $F^i$ are different from the current solution is found. 
This modification is called \emph{Exhaustive Donor Search (EDS)}, following \cite{goldman2015fast}.

\subsubsection{Forced Improvements (FI)}
If no subset $F^i$ leads to changes in the solution undergoing GOM, the so-called \emph{forced improvements} (FI) phase can (optionally) start.
Originally, the FI was proposed in \cite{bosman2012linkage} to deal with convergence issues in MAXCUT. Namely, it can happen that the population starts to drift in fitness plateaus, i.e., solutions keep changing without improving. This lack of convergence makes it unlikely for further improvements to be discovered.
Therefore, FI is specifically designed  to steer the search towards converging to the (or one, if there is many) elitist solution.
The FI phase works like the normal GOM phase, except for the fact that the donor solution is always set to be the elitist solution.
Moreover, to further ensure convergence, changes that lead to equal fitness are now rejected (one can no longer drift in fitness plateaus). 
Only if the solution strictly improves in fitness, the overwrite action is accepted. To prevent the FI phase to reduce diversity too fast, the FI phase is stopped as soon as an improvement happens. 
Finally, if a solution could not be improved in the FI phase, it is overwritten by the elitist solution. 

\subsection{Conditional Gene-pool Optimal Mixing (CGOM)} 
By design, the GOM operator copies genes from a donor solution independently for each FOS element. 
Therefore, dependencies between FOS elements are not taken into account, i.e., when GOM is applied to a FOS element, any (weak) dependencies of variables inside the FOS element to variables outside the FOS element are not considered which might lead to suboptimal linkage usage because it may well be that although interactions between variables are of low order, they may still not be defined in terms of mutual exclusive subsets. I.e., consider the NK-landscapes with random subsets of variables for the subfunctions
To alleviate this limitation, we consider a new gene-pool optimal mixing operator - the \emph{Conditional GOM (CGOM)}.
CGOM is closely related and inspired by recently introduced conditional linkage models for the real-valued GOMEA \cite{bouter2020leveraging}. However, in RV-GOMEA conditional dependencies were not considered together with a hierarchical model like the LT which we do have for the first time. 

CGOM works similar to GOM but takes into account what gene values are being processed to choose suitable donor solutions. Specifically, each FOS subset can be made conditionally dependent on a group of other variables.
If the variables contained in a FOS subset $F^i$ are conditionally dependent on variables not in $F^i$, CGOM takes this into account during mixing. 
Suppose some genes have already been considered during mixing, i.e., for the current application of GOM to a given solution, these variables have been subjected to GOM before (they were in a FOS element considered earlier). We store these genes in a set $U$. 
When a new FOS element $F^i$ is considered, we compute (explained below) the set of variables $V ~\textnormal{s.t.}~ 1)~V \cap F^i = \emptyset$, 2) all variables in $V$ depend on the variables in $F^i$ (we refer to such set of variables as $G_i$), and 3) $V \subseteq U$ (i.e., they were considered before). 
Since variables from $V$ and $F^i$ are interdependent, we enforce that selecting which genes configuration for $F^i$ is considered, should be conditioned on $V$. 
This is achieved by considering as donor solutions only those which have the same genes for variables in $V$ as the current solution undergoing CGOM has. 

\begin{figure*}
    \centering
    \includegraphics[width=0.95\textwidth]{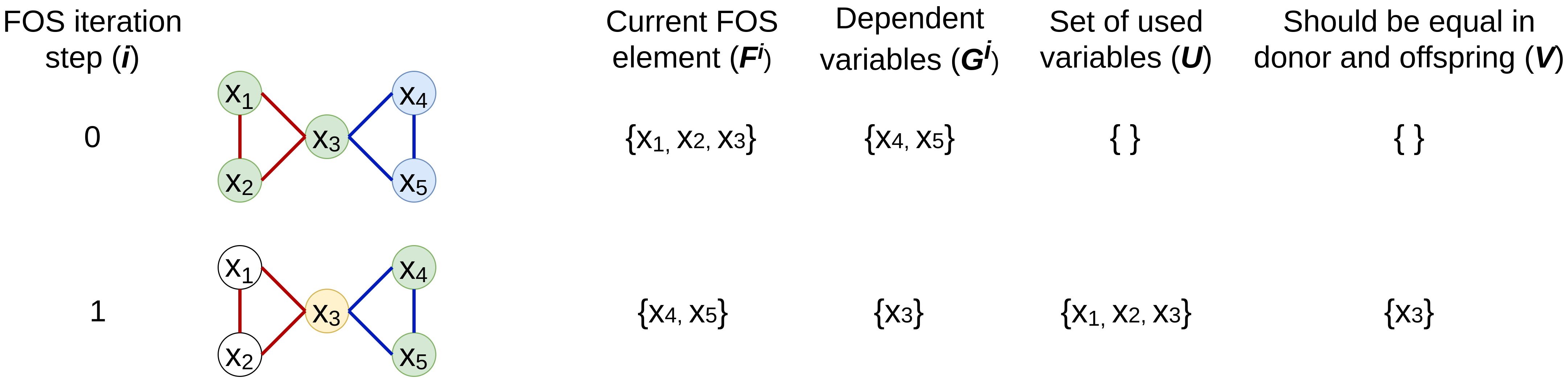}
    \caption{
    A minimal CGOM working example. Fitness function $f$ is a function of 5 variables and can be decomposed into two subfunctions: $f(x_1, x_2, x_3, x_4, x_5)=f_1(x_1, x_2, x_3)+f_2(x_3, x_4, x_5)$. This is shown by the different colors of edges. Suppose that after filtering the LT FOS contains two elements: $\{x_1, x_2, x_3\}$ and $\{x_4, x_5\}$. Variables from the current FOS element are colored in green; variables which are dependent with it are colored in blue; variables which are dependent with it and are already used (i.e., are taken into account by CGOM) are colored in yellow. Green colored genes are pooled to an offspring from only those donors which have the yellow colored genes equal to what is found in the current offspring.}
    \label{fig:CGOM_example}
\end{figure*}

A minimal CGOM working example is shown in Figure \ref{fig:CGOM_example}. 
The CGOM differences as compared to GOM in terms of pseudocode are highlighted in Algorithm~\ref{alg:GOM}. 

In the BBO paradigm, we have no a priori information on the dependence structure between variables. 
However, similar to FOS learning, we can estimate a notion of variable dependence based on the state of the population and the similarity measure (e.g., MI or NMI). 
Broadly speaking, we say that a FOS element $F^i$ is dependent on a variable $X$ ($X \notin F^i$) if the average pairwise similarity measure between the variables in $F^i$ and $X$ is relatively large compared to the average similarity measure between the variables only in $F^i$ on the one hand and all the variables that do not belong to $F^i$ on the other hand. 
Particularly, we use a threshold to detect such dependencies: a FOS element $F^i$ is considered to have a dependency with variable $j$ if the average pairwise similarity measure between $j$ and variable in $F^i$ is greater than $\lambda M$ where $M$ is the largest average pairwise similarity score between variables from $F^i$ and variables not belonging to $F^i$. 
This dependencies learning procedure is described in pseudocode in the function \emph{learnDependencies} of Algorithm~\ref{alg:singlepop}. 
The Hyperparameter $\lambda$ is tunable. The smaller the value of $\lambda$, the larger the number of estimated dependencies. In other words, small $\lambda$ values will result in high recall (we are unlikely to miss dependencies but might have many false positives), while large $\lambda$ values will improve precision (we might miss many dependencies but will have few small positives).

 \begin{algorithm}
 \def\HiLi{\leavevmode\rlap{\hbox to \hsize{\color{yellow!50}\leaders\hrule height .8\baselineskip depth .5ex\hfill}}}

 \DontPrintSemicolon
\footnotesize
 \SetKwFunction{evaluate}{evaluate}
 \SetKwFunction{cp}{copy}
 \SetKwFunction{CGOM}{CGOM}
 \SetKwFunction{randomlySelect}{randomlySelect}
 \SetKwFunction{randomPermutation}{randomPermutation}
 \SetKwFunction{evaluateAndUpdateElitist}{evaluateAndUpdateElitist}
 \SetKwFunction{evaluateFitness}{evaluateFitness}
 \SetKwFunction{acceptChange}{acceptChange}
 \SetKwFunction{FI}{FI}
 \SetKwFunction{checkDonor}{checkDonor}
 
\SetKwProg{Fn}{Function}{:}{}
\SetKwData{true}{true}
\SetKwData{false}{false}
\newcommand\mycommfont[1]{\footnotesize\ttfamily\textcolor{blue}{#1}}
\SetCommentSty{mycommfont}
\Fn{\CGOM{s, $\mathcal{G}$, useEDS, useFI}}
{
$b \gets o \gets s$ \;
$changed \gets \false$ \;
\HiLi{$U \gets \emptyset$} \;
$\mathcal{F} \gets orderFOS(\mathcal{F})$\;
\For{$i \in \{0,1,\dots,|\mathcal{F}|-1\}$}
    {
        $donorsList = \randomPermutation (\{\mathcal{P}_0, \mathcal{P}_1,\dots,\mathcal{P}_{n-1}\})$\;
        \For{$j \in \{0,1,\dots,n-1\}$}
        {
            \HiLi{$U \gets U \cup F^i$ \;}
            $d \gets donorsList[j]$\;
            \HiLi{\If{$\neg \checkDonor(o, d, F^i, \mathcal{G}, U)$}
            {
                \HiLi{continue\;}
            }
            }
            $o_{F^i} \gets d_{F^i}$\;
            \If {$o_{F^i} \ne d_{F^i}$}
                {
                    $\evaluateAndUpdateElitist(o)$\;
                    \If{\acceptChange(o)}
                    {
                        $b_{F^i} \gets o_{F^i}$\;
                        $changed \gets \true$ \;
                    }
                    \Else
                    {
                        $o_{F^i} \gets b_{F^i}$\;
                    }

                    break\;
                }
            \If{$\neg useEDS$}{break\;}
        }   
    }
    \If{$useFI~ and (\neg changed~ or~ o.NIS > 1 + \log_{10}(n)$)}
    {
        \HiLi{\FI(s)};
    }
    \If{$o.fitness \le s.fitness$}
    {
        $o.NIS \gets o.NIS+1$\;
    }
    \Else
    { 
        $o.NIS \gets 0$\;
    }
}

\Fn{\checkDonor{$o, d, F^i, \mathcal{G}, U$}}
{
    
       \tcc{$\mathcal{G}(v)$ are variables linked with $F^i$}
        \For{$p \in \mathcal{G}(v) \cap U$}
        {
            \If{$o.v \ne d.v$}
            {
                return \false;
            }
        }
    return \true;
}

\Fn{\FI(s)}
{
$changed \gets \false$ \;
\HiLi{$U \gets \emptyset$ \;}
$\mathcal{F} \gets orderFOS(\mathcal{F})$\;
\For{$i \in \{0,1,\dots,|\mathcal{F}|-1\} $}
{
    \HiLi{\If{$\neg \checkDonor(o, elitist, F^i, \mathcal{G}, U)$}
    {
        \HiLi{continue\;}
    }}
    \HiLi{$U \gets U \cup F^i$ \;}
    \If {$o_{F^i} \ne elitist_{F^i}$}
    {

        $o_{F^i} \gets elitist_{F^i}$\;
        \evaluateAndUpdateElitist(o)\;
        \If{$o.fitness > b.fitness$}
        {
            $changed \gets \true$ \;
            break\;
        }
        \Else
        {
            $o_{F^i} \gets b_{F^i}$\;
        }
    }
\If{$\neg useEDS$}{break\;}

}
\If {$\neg changed$}
{
    $o \gets elitist$ \;
}
}

\caption{GOM and the proposed CGOM operators. Highlighted lines are used in CGOM only. 
\label{alg:GOM}}
\end{algorithm}

\subsection{GOMEA with a Traditional, Single Population} \label{subsection:singlePop}
The initial population of $n$ solutions is initialized randomly.
After random solutions are generated, a local search algorithm can be applied to efficiently move them to a local optimum. 

\subsubsection*{Local Search}
We consider two local search algorithms here: simple \emph{Single-Iteration Hill Climber} (\emph{SIHC}) and \emph{Exhaustive Hill Climber} (\emph{EHC}) which is SIHC repeated multiple times until no improvements are found in a single iteration over all variables. 
Both hill climber variants were shown to be efficient components of advanced EAs, for instance, SIHC was used in \cite{dsmga2}, and EHC was used in \cite{goldman2015fast}. 
The pseudocode for considered Hill Climber algorithms is listed in Algorithm \ref{alg:HC}.

\subsubsection*{Tournament Selection}
Each iteration of the main GOMEA loop starts with linkage model learning. 
GOMEA does not have a traditional selection phase because GOM already induces selection, by discarding changes that are detrimental to a solution's fitness. 
However, we consider the option of using tournament selection to select good solutions upon which to learn the linkage model, as done in \cite{dsmga2, chen2017two}. 
We remark that with this option, the selection is disregarded after the linkage model is learned, i.e., it is not used to override the population.

After the linkage model is learned, the GOM variation operator is applied to every solution in the population to generate $n$ offspring solutions. The population is then completely replaced
by the offspring solutions. This main loop runs until the termination criterion
is satisfied, which is naturally triggered when the population converges (i.e., all solutions have equal genotypes), but can also include other termination conditions such as a maximum allowed runtime, a maximum number of function evaluations, or a maximum number of generations. 
The pseudocode of single-population GOMEA is provided in Algorithm \ref{alg:singlepop}.
 \begin{algorithm}
 \DontPrintSemicolon
\footnotesize
 \SetKwFunction{randomPermutation}{randomPermutation}
 \SetKwFunction{evaluate}{evaluate}
 \SetKwFunction{randomlyInitialize}{randomlyInitialize}
 \SetKwFunction{cp}{copy}
 \SetKwFunction{termination}{TerminationCriterionSatisfied}
 \SetKwFunction{HC}{exhaustiveHillClimber}
 \SetKwFunction{singleHC}{singleIterationHillClimber}
\SetKwProg{Fn}{Function}{:}{}
\SetKwData{true}{true}
\SetKwData{false}{false}
\SetKwRepeat{Do}{do}{while}
 \Fn{\HC{s}}
 {
 \Do{$improved$}
 {
    $s, improved \gets$ \singleHC{s} \;
 }
 }
 
 \Fn{\singleHC{s}}
 {
 $improved = \false$ \;
 $f_s \gets$ \evaluate{$s$}\;
 \For{$i \in$ \randomPermutation{$\{0, \dots, \ell-1\}$}}
 {
     $s^\prime \gets$ \cp{$s$} \;
     
     $ s^{\prime}_i \gets s^{\prime}_i \oplus 1 $\;
     $f_{s^\prime} \gets$ \evaluate{$ s^\prime $} \;
     \If{$f_{s^\prime} > f_s$}
     {
         $s_i \gets s^{\prime}_i$\;
         $f_s \gets f_{s^\prime}$\;
         $improved \gets \true$ \;
     }
 }
 return $s, improved$
}
\caption{Single-Iteration and Exhaustive Hill Climber algorithms . 
\label{alg:HC}}
\end{algorithm}

\begin{algorithm}
 \def\HiLi{\leavevmode\rlap{\hbox to \hsize{\color{yellow!50}\leaders\hrule height .8\baselineskip depth .5ex\hfill}}}

 \DontPrintSemicolon
\footnotesize
 \SetKwFunction{evaluate}{evaluate}
 \SetKwFunction{cp}{copy}
 \SetKwFunction{singlePopulationGOMEA}{singlePopulationGOMEA}
 \SetKwFunction{createPopulation}{createPopulation}
 \SetKwFunction{randomlySelect}{randomlySelect}
 \SetKwFunction{learnModel}{learnModel}
 \SetKwFunction{randomPermutation}{randomPermutation}
 \SetKwFunction{doOneGeneration}{doOneGeneration}
 \SetKwFunction{evaluateAndUpdateElitist}{evaluateAndUpdateElitist}
 \SetKwFunction{evaluateFitness}{evaluateFitness}
 \SetKwFunction{createRandomSolution}{createRandomSolution}
 \SetKwFunction{hillClimber}{hillClimber}
 \SetKwFunction{learnDependencies}{learnDependencies}
 \SetKwFunction{calculateSimilarityMatrix}{calculateSimilarityMatrix}
 \SetKwFunction{max}{max}
 
\SetKwProg{Fn}{Function}{:}{}
\SetKwData{true}{true}
\SetKwData{false}{false}
\newcommand\mycommfont[1]{\footnotesize\ttfamily\textcolor{blue}{#1}}
\SetCommentSty{mycommfont}
\Fn{\singlePopulationGOMEA{useHC, useEDS, useFI}}
{
\While{$\neg terminationCriterionSatisfied$}
{
    $\mathcal{P} \gets \doOneGeneration(\mathcal{P})$ \;
}
}

\Fn{\createPopulation{n, useHC}}
{
\For{$i \in \{0,1,\dots,n-1\}$}
{
    $\mathcal{P}_i \gets \createRandomSolution()$ \;
    \If {useHC}
    {
    $\mathcal{P}_i = \hillClimber(\mathcal{P}_i)$ \;
    }
    $\evaluateAndUpdateElitist(\mathcal{P}_i)$ \;
}
}

\Fn{\doOneGeneration{$\mathcal{P}$}}
{
    $\mathcal{F} \gets \learnModel(\mathcal{P})$ \;
    $\mathcal{F} \gets \mathcal{F}\setminus\{\{0,1,\dots,\ell-1\}\}$\;
    \HiLi{$\mathcal{G} \gets \learnDependencies(\mathcal{F}, \mathcal{P})$} \;
    \For{$i \in \{0,1,\dots,n-1\}$}
    {
        $\mathcal{O}_i \gets GOM(\mathcal{P}_i, useEDS, useFI)$\;
        \tcc{GOM usage can be changed to CGOM}
        \HiLi{$\mathcal{O}_i \gets CGOM(\mathcal{P}_i, \mathcal{G}, useEDS, useFI)$\;}
    }
    $\mathcal{P} \gets \mathcal{O}$ \;
    return $\mathcal{P}$ \;
}

\Fn{\learnDependencies{$\mathcal{F},\mathcal{P}$}}
{
$\mathcal{G} \gets \{\emptyset\}_{i=0}^{|\mathcal{F}|-1}$ \; 
$\mathcal{S} \gets \calculateSimilarityMatrix(\mathcal{P})$ \;
\For{$i \in \{0,1,\dots,\mathcal{F}-1\}$}
{
    $R \gets \{0\}_{j=0}^{\ell-1}$  \; 
    \For{$j \in \{0,1,\dots,\ell-1\} \setminus F^i$}
    {
        $R_j \gets \frac{1}{|F^i|} \sum_{k=0}^{|F^i|-1}S_{j,F^i_k}$ \;
    }
    $M \gets \max(R)$ \;
    \For {$j \in \{0,1,\dots,\ell-1\} \setminus F^i$}
    {
        \If{$R_j > \lambda M > 0$}
        {
            $\mathcal{G}^i \gets \mathcal{G}^i \cup \{j\}$ \;
        }
    }
}
return $\mathcal{G} \;$
}

\caption{Single population GOMEA. Necessary modifications to use CGOM instead of GOM are highlighted.
\label{alg:singlepop}}
\end{algorithm}

\subsection{Going parameterless: removing the need to set the population size}
The population size is a crucial parameter for the success of EAs. With model-based EAs like GOMEA, this is arguably even more so because the linkage model needs sufficient samples to be learned to achieve a sufficient level of accuracy for the linkage to be reliable. However, choosing the right population size is problem-dependent, and highly non-trivial. Methods to scale the population size automatically over time are therefore extremely useful and convenient in practice. 
In this paper, we consider two well-known population size-free schemes.

First, we consider the Interleaved Multistart Scheme (IMS), which was heavily inspired from the work by Harik and Lobo on parameterless GAs \cite{harik1999parameter}. The IMS has been shown to be easy-to-use and can be naturally applied to almost any EA in various optimization domains \cite{luong2018multi, lin2018investigation, dushatskiy2019convolutional, GP_IMS}.

The IMS consists of evolving multiple populations simultaneously, in an interleaved fashion. Its pseudocode with recursive implementation is listed in Algorithm \ref{alg:IMS}. In the beginning, a single population is initialized, typically of a very small size (e.g., 2). After $\mathcal{M}_{IMS}$ generations, a new population is initialized that is larger, and it is advanced by one generation. This larger population will execute its next generation only after the smaller population performs $\mathcal{M}_{IMS}$ generations more. When the larger population has performed $\mathcal{M}_{IMS}$ generations, an even larger population is initialized, and so on. 
Our implementation of the IMS uses an initial population of size $2$, exponential growth whereby each new population is twice the size of the previous, and $\mathcal{M}_{IMS}=4$, as in \cite{harik1999parameter}.
Smaller populations are terminated if they have converged, or their average fitness has become smaller than the average fitness of a larger population. 
This is because when a larger population has caught up with the smaller population, the latter will likely converge sooner, and can therefore be considered obsolete. 
Additionally, another convergence criterion such as a maximum allowed number of generations per population can be implemented.

\begin{algorithm}
 \def\HiLi{\leavevmode\rlap{\hbox to \hsize{\color{yellow!50}\leaders\hrule height .8\baselineskip depth .5ex\hfill}}}

 \DontPrintSemicolon
\footnotesize
 \SetKwFunction{evaluate}{evaluate}
 \SetKwFunction{cp}{copy}
 \SetKwFunction{IMS}{IMS}
 \SetKwFunction{doOneGeneration}{doOneGeneration}
 \SetKwFunction{checkTermination}{checkTermination}
 \SetKwFunction{generationalStep}{generationalStep}
 \SetKwFunction{createRandomSolution}{createRandomSolution}
 \SetKwFunction{averageFitness}{averageFitness}
 \SetKwFunction{converged}{converged}

\SetKwProg{Fn}{Function}{:}{}
\SetKwData{true}{true}
\SetKwData{false}{false}
\newcommand\mycommfont[1]{\footnotesize\ttfamily\textcolor{blue}{#1}}
\SetCommentSty{mycommfont}
\Fn{\IMS{useHC, useEDS, useFI}}
{
$Populations \gets []$ \;
\While{$\neg terminationCriterionSatisfied$}
{
    $\mathcal{P} \gets \createPopulation(2^{|Populations|+1}))$ \;
    $Populations.append(\mathcal{P})$ \;
    $\generationalStep(Populations)$ \;
}
}

\Fn{\checkTermination{$Populations, i$}}
{
    \If {$\converged(Populations^i)$}
    {
    return \true\;
    }
    \For {$j \in \{i+1,\dots,|Populations|-1\}$}
    {
        \If {$\averageFitness(Populations^j) > \averageFitness(Populations^i)$}
        {
            return \true;
        }
    }
    return \false;
}

\Fn{$\generationalStep{Populations, first, last}$}
{
\For {$iter \gets \{0,1,\dots,\mathcal{M}_{IMS}-1\}$}
{
    \For {$i \in \{first,\dots,last\}$}
    {
        \If {$\neg Populations^i.terminated$}
        {
            $Populations^i.terminated \gets \checkTermination(Populations, i)$ \;
        }
        \If {$\neg Populations^i.terminated$}
        {
            \doOneGeneration{$Populations^i$} \;
        }
    }
    \For {$i \in \{first,\dots,last-1\}$}
    {
        \generationalStep{$Populations, first, i$} \;
    }
}
}

\caption{Interleaved Multistart Scheme (IMS).
\label{alg:IMS}}
\end{algorithm}

The other schemes that we consider are the Population Pyramid (P3) \cite{goldman2015fast}, and its further modification Multiple Insertion Pyramid (P3-MI) \cite{den2016multiple}. The difference between the two is explained below and pseudocode is provided in Algorithm \ref{alg:MIP3}. 

P3-MI arranges the population into a pyramidal structure, whereby each level of the pyramid is a set of solutions (duplicates are not stored).
When a new population is created and, optionally, a local search is applied, all solutions are added to the bottom level of the pyramid. 
Then, by using solutions from the current pyramid level as donors, offsprings of the current population are generated. 
Solutions that are improved by variation are promoted and entered into one level higher in the pyramid.
If there is no next level in the pyramid, a new one is created. 
This process continues until the pyramid's top level is reached or no solutions are improved during a generation. Every generation, Linkage models are learned for each pyramid layer independently. Sizes of populations are determined by a population \emph{growth function}. The growth function takes the iteration number as input, and produces the population size (i.e., the number of solutions added to the bottom level of the pyramid).
In \cite{goldman2015fast} different population growth functions were studied. In this work we use a quadratic function ($t^2$, where $t$ is the iteration, starting from 1) as a trade-off between speed and number of function evaluations. 

The P3 scheme is a special case of P3-MI with a constant growth function with value $1$, in other words, one new solution is created and evolved in each iteration. 

\begin{algorithm}
 \def\HiLi{\leavevmode\rlap{\hbox to \hsize{\color{yellow!50}\leaders\hrule height .8\baselineskip depth .5ex\hfill}}}

 \DontPrintSemicolon
\footnotesize
 \SetKwFunction{evaluate}{evaluate}
 \SetKwFunction{cp}{copy}
 \SetKwFunction{MIPGOMEA}{P3MI}
 \SetKwFunction{growthFunction}{growthFunction}
 \SetKwFunction{createRandomSolution}{createRandomSolution}
 \SetKwFunction{hillClimber}{hillClimber}

\SetKwProg{Fn}{Function}{:}{}
\SetKwData{true}{true}
\SetKwData{false}{false}
\newcommand\mycommfont[1]{\footnotesize\ttfamily\textcolor{blue}{#1}}
\SetCommentSty{mycommfont}
\Fn{\MIPGOMEA{useHC, useEDS, useFI}}
{
$iter \gets 0$ \;
$Pyramid \gets [\emptyset$] \;
\While{$\neg terminationCriterionSatisfied$}
    {
        $n \gets \growthFunction(iter)$ \; 
        $\createPopulation(n, useHC)$ \;
        $Pyramid^0 \gets Pyramid^0 \cup \mathcal{P}$ \;
        $solutionsAdded \gets \true$ \;
        $currentTopLevel \gets |Pyramid|-1$ \;
        $\mathcal{L} \gets 0$ \;
        \While {$\mathcal{L} \le currentTopLevel$ and $solutionsAdded$}
        {
            $\mathcal{F} \gets \learnModel(Pyramid^\mathcal{L})$ \;
            $\mathcal{F} \gets \mathcal{F}/\{\{0,1,\dots,\ell-1\}\}$ \;
            \For{$i \in \{0,1,\dots,n-1\}$}
            {
                $\mathcal{O}_i \gets GOM(\mathcal{P}_i, Pyramid^\mathcal{L})$ \;
                \If {$\mathcal{O}_i.fitness > \mathcal{P}_i.fitness$}
                {
                    \If{$\mathcal{L}=currentTopLevel$}
                    {
                        $Pyramid^{\mathcal{L}+1}.append(\emptyset)$ \;
                    }
                    $Pyramid^{\mathcal{L}+1} \gets Pyramid^{\mathcal{L}+1} \cup \{\mathcal{O}_i\}$ \;
                    $solutionsAdded \gets \true$ \;
                }
            }
            $\mathcal{P} \gets \mathcal{O}$ \;
            $\mathcal{L} \gets \mathcal{L}+1$ \;
        }
        
        $iter \gets iter+1$ \;
    }
}

\caption{P3-MI population management scheme. P3 is a special of case P3-MI when population growth function is constant and has value 1 for all iterations.
\label{alg:MIP3}}
\end{algorithm}

\section{Experiments}\label{sec:sectionexperiments}
\subsection{Benchmark problems}
We consider various combinatorial optimization problems that are commonly considered to be particularly interesting for benchmarking GAs.

\subsubsection{Concatenated deceptive traps}
Concatenated deceptive trap is a well-known benchmark problem that was introduced to show that with disrupting building blocks, it takes exponentially growing resources to solve this problem. The fitness function of this problem is defined as:
$$f_{Trap_K^S}(x) = \sum\limits_{i \in \{0,s,2s,...\},i < \ell }f^{sub}_{TrapK}\left(\sum\limits_{j=0}^{k-1}x_{(i+j) \% \ell}\right) $$ $$f^{sub}_{TrapK}(u)=\begin{cases} k & \mbox{if~} \mbox{$u=k$} \\ {k-1-u} & \mbox{otherwise} \end{cases}$$
Particularly, we consider trap functions with subfunctions size $k=5$ and two different values of subfunctions overlap: separable traps with $s=5$ (further referred to as $Trap_5^5$) and overlapping traps with $s=4$ ($Trap_5^4$).

\subsubsection{Bimodal separable deceptive trap}
The bimodal symmetric concatenated trap functions \cite{deb1993multimodal} are interesting because in contrast to standard concatenated trap described above each subfunction has two modes. We consider bimodal symmetric traps of size 6, such that the subfunctions are given by
$$f^{sub}_{BimodalTrapK}(u)=
\begin{cases} 
6 & \mbox{if~} \mbox{$u=0$ or $u=6$} \\ 
0 & \mbox{if~} \mbox{$u=1$ or $u=5$} \\
2 & \mbox{if~} \mbox{$u=2$ or $u=4$} \\
5 & \mbox{otherwise} 
\end{cases}$$

The considered bimodal traps do not have subfunctions overlap.

\subsubsection{NK-landscapes}
The NK-landscapes with maximum overlap (also called NK-S1 landscapes) \cite{nk-landscapes} with subfunctions of size $k=5$ are interesting because of overlapping subfunctions which are different depending on the position in genotype.
$$f_{NK}(x) = \sum\limits_{i=0}^{l-k}f_{NK}^{sub}(x_{(i,i+1\,\dots,i+k)}$$ where the values of $f_{NK}^{sub}$ are tabular values, sampled from the uniform distribution in $[0;1]$ interval independently for different subfunctions positions.

\subsubsection{Hierarchial-If-And-Only-If (HIFF)}
The Hierarchical If-And-Only-If (HIFF) function is interesting because includes hierarchically ordered dependencies of exponentially growing sizes that overlap:
$$f_{Hiff}(x) = \sum\limits_{k\in\{1, 2, 4 \dots \frac{l}{2},l\} {}}\sum\limits_{i=0}^{l/k-1}f^{sub}_{Hiff}(x_{ik \dots (i+1)k-1})$$  $$f^{sub}_{Hiff}(u)=\begin{cases} 1 & \mbox{if } \sum\limits_{j=0}^{k-1}{}u_j=k \mbox{ or } \sum\limits_{j=0}^{k-1}u_j=0 \\ 0 & \mbox{otherwise} \end{cases}$$

\subsubsection{MAXCUT}
We consider MAXCUT as a well-known combinatorial optimization problem. Given a weighted undirected graph ($V, E$) the goal is to find a partition of the vertices in two sets such that the sum of weights of edges running between vertices in different partitions is maximized. The fitness function is therefore defined as:
$$f_{MAXCUT}=\sum_{(i,j) \in E: x_i \ne x_j}{w_{ij}}$$ where $w_{ij}$ is the weight of edge $(i,j)$.

We consider two types of MAXCUT instances. The first type is 3D square torus graphs. Each vertex is connected to 4 neighbors, forming a torus. Edges weights are integer values from $[1, 5]$ sampled uniformly. This type of instance is further referred to as MAXCUT Sparse. The second type of instance is dense graphs with randomly selected $\sqrt{\ell}$ neighbors for each vertex. This type of MAXCUT instances further referred to as MAXCUT Dense, and they are known to be NP-hard problem.
For MAXCUT Dense, we use edge weights values in $[0,1000]$ sampled uniformly.

\subsubsection{Ising spin-glass}
2D Ising spin-glass problems have often been considered in the benchmarking of EDAs and other model-based EAs.
The spin-glass problem fitness function is defined as 
$$
f_{spinglass}(x) = \sum_{i=0}^{\ell-1} \sum_{j=0}^{\ell-1} x_i x_j J_{ij}
$$
where $J_{ij}$ defines an interaction value between two variables, $J_{ij} \in \{-1,1\}$. In the considered spin-glass instances each variable interacts with up to 4 neighbors in a 2D grid.

\subsubsection{MAXSAT}
Finally, we consider is MAXSAT problem. Particularly, we consider unweighted MAX-3SAT uniform random instances \cite{chen2017two}. MAX-3SAT is NP-hard.

$$
f_{MAXSAT}(x) = \sum_{i=0}^{m-1} (\lor_{j=0}^{p_i-1} \Gamma_{ij} x_{f_{ij}}) $$

where $m$ is the number of subfunctions (clauses), $p_i$ is subfunction size (in the used instances $~\forall i~ p_i=3$), $f_i$ determines which variables are contained by the subfunction with index $i$, and $\Gamma$ can be either an unary negation operator (turning a binary $x$ to an opposite value) or an identity operator keeping the value of $x$ intact.

\subsection{Sizes of problems}
We frame our experiments in terms of scalability, i.e., we record what the effort is (in terms of time and function evaluations) for an EA to find the optimum, for growing problem dimensionality.
For experiments where we traditionally adopt a single population, the maximum dimensionality we consider is set to 640 for $\mathrm{Trap_5^5}$, $\mathrm{Trap_5^4}$, and NK-S1, to 636 for Bimodal Trap, to 1600 for MAXCUT Sparse, to 784 for Spin-glass, to 1024 for HIFF, and to 100 for NP-hard MAXCUT Dense and MAXSAT. 
In experiments with automatic population sizing schemes, the maximum problem sizes are doubled for all problems except for MAXCUT Sparse and Spin-glass. 
For the experiments with a single population we need to use smaller maximal dimensionalities because we included bisection to discover what the optimal population size is, but bisection quickly becomes computationally prohibitive to run for large problems.

\newcommand{\gomeabest}{\emph{GOMEA}\textsuperscript{\emph{best}~}}
\newcommand{\cgomeabest}{\emph{CGOMEA}\textsuperscript{\emph{best}~}}
\newcommand{\gomeabestp}{\emph{GOMEA-P3}\textsuperscript{\emph{best}~}}
\newcommand{\cgomeabestp}{\emph{CGOMEA-P3}\textsuperscript{\emph{best}~}}

\subsection{Finding the best settings for single-population GOMEA}\label{hyperparamsGOM}
We summarize different possible choices of single-population GOMEA components in Table \ref{tab:params}.
Since we are interested in eliminating the need to choose parameters, we attempt to define what the best GOMEA variant is across the different benchmark problems.

\begin{table}[ht]
\caption{Considered hyperparameters of single-population GOMEA. In bold, the best settings found by the experiment described in Section~\ref{hyperparamsGOM}.}
\label{tab:params}
\centering
  \begin{tabularx}{0.45\textwidth}{|l|X|}
  \hline
    \textbf{Hyperparameter}  & Options \\
    \hline
    Forced Improvements       & \textbf{on} / off \\
    Exhaustive Donor Search   & \textbf{on} / off \\
    Hill Climber              & \textbf{SIHC} / EHC / off \\
    Linkage Tree and similarity measure & unfiltered, MI / \textbf{filtered, NMI}\\
    FOS ordering           & random / \textbf{ascending subsets size} \\
    Tournament Selection (size 2)       & \textbf{on} / off \\
  \hline
  \end{tabularx}
\end{table}

In total, there are 96 ($2^5*3$) combinations of hyperparameters.
We perform an exhaustive hyperparameter search by running all 96 GOMEA variants on a set of benchmark problems. 
Here, our goal is to fairly compare all GOMEA variants. 
In order to do so, for the largest considered size of each problem, we carry out the comparisons among configurations that all have a respective optimal population size. 
We estimate the optimal population size using the bisection method.
The success condition in bisection is solving (i.e., achieving a global optimum) a problem instance in every of 50 consecutive runs. 
We do not put any hard constraints on runtime. Instead, we bound it by limiting the total number of function evaluations by $10^8$ and, additionally, the total number of generations of each population by 200 (the same value as used in \cite{chen2017two}) to prevent convergence problems. 
As it might happen that the smallest population size which allows to solve a problem instance does not require the fewest function evaluations, during the bisection procedure we keep track of the population size which allows to solve a problem instance with the fewest function evaluations. 
If the population size reaches $10^5$ solutions and a problem instance is still not solved, the optimal population search procedure is terminated.

We rank the variants of GOMEA based on the minimal number of evaluations taken to find the optimal solution for each problem. 
The final ranking of a variant is the average of the rankings across the problems. 
If a variant is not able to solve one or more problems, it is dropped from the comparison.

The best GOMEA version is further referred to as \gomeabest.

\subsection{Adding CGOM operator}
Once \gomeabest is found, we look into the effect of replacing GOM with the new CGOM operator. Since CGOM requires a detection threhsold $\lambda$ to be set, we run comparisons with $\lambda \in \{0.5, 0.6, 0.7, 0.8, 0.9\}$. 
We determine the best performing value of $\lambda$ using the same approach as in Section \ref{hyperparamsGOM}. 
This best performing CGOMEA version is further referred to as \cgomeabest.

\subsection{Benchmarking algorithms using the optimal population size}
\cgomeabest and \gomeabest are compared against each other, against the best previously published version of GOMEA \cite{thierens2011optimal}, and against the most recent single-population DSMGA-II version \cite{chen2017two}. The optimal population sizes for all algorithms are determined using bisection. To test the statistical significance of performance differences between two algorithms we use the pairwise Mann-Whitney $U$-test. 
For each problem we do a separate test which checks the hypothesis that the first algorithm needs fewer evaluations than the second algorithm to solve this problem in its maximum dimensionality. 

\subsection{Finding the best settings for parameterless algorithms} \label{hyperparamsParameterlessGOM}
Next, we find the best performing parameterless version of GOMEA. 
The considered options of a parameterless scheme are IMS, P3-MI with quadratic population growth function, and P3. The scheme is seen as another tunable hyperparameter. We combine it with 96 hyperparameter combinations described in Section~\ref{hyperparamsGOM} and perform a large-scale hyperparameter search, consisting of 96*3=288 possible algorithm configurations. This best performing parameterless GOMEA version is further referred to as \gomeabestp.

Once \gomeabestp is found, we replace GOM with the new CGOM operator (with $\lambda$ value which was chosen for \cgomeabest, i.e., 0.8). This CGOMEA version is further referred to as \cgomeabestp.

\begin{table}[ht]
\caption{Considered hyperparameters of parameterless GOMEA. In bold, the best settings found by the experiment described in Section~\ref{hyperparamsParameterlessGOM}.}
\label{tab:hyperparams_popfree}
\centering
  \begin{tabularx}{0.45\textwidth}{|l|X|}
  \hline
    \textbf{Hyperparameter}  & Options \\
    \hline
    Forced Improvements       & on / \textbf{off} \\
    Exhaustive Donor Search   & \textbf{on} / off \\
    Hill Climber              & \textbf{SIHC} / EHC / off \\
    Linkage Tree and similarity measure & unfiltered, MI / \textbf{filtered, NMI}\\
    FOS ordering           & \textbf{random} / ascending subsets size \\
    Tournament Selection (size 2)       & on / \textbf{off} \\
    Population scheme       & \textbf{P3} / P3-MI / IMS \\
  \hline
  \end{tabularx}
\end{table}

Additionally, we add to the experiments the original P3 algorithm and DSMGA-II with IMS \cite{lin2018investigation}. 
Note that we do not test other population management schemes for DSMGA-II since, to the best of our knowledge, their integration with DSMGA-II have not been studied. 

To study the practical applicability of the algorithms, we remove the limit on the number of function evaluations. Instead, in all experiments with parameterless algorithms we set a time limit of 24 hours.
This is needed to make experiments computationally feasible as some of the considered algorithms (especially some configurations which use the P3 scheme and DSMGA-II with IMS) perform in a way that the number of function evaluations is increasing very slowly.

\subsection{Implementation details}
All GOMEA variants and the P3 algorithm are implemented in C++\footnote{Source code is available on the repository of the first author: \url{{https://github.com/ArkadiyD/BinaryGOMEA}} and the website of the last author: \url{http://www.cwi.nl/~bosman}}. 
The P3\footnote{\url{https://github.com/brianwgoldman/FastEfficientP3/}} and DSMGA-II\footnote{\url{https://github.com/tianliyu/DSMGA-II-TwoEdge}} implementations are the ones used in their corresponding original articles with modified fitness functions
to make them identical for all conducted experiments. Compiler settings for all considered algorithms are also identical.

\section{Results}\label{sec:Results}

\newcommand{\problema}{\textbf{\textit{Trap}$\mathrm{\boldsymbol{{}_5^5}}$}}
\newcommand{\problemb}{\textbf{\textit{Trap}$\mathrm{\boldsymbol{{}_5^4}}$}}
\newcommand{\problemc}{\textbf{\textit{Bimodal Trap}}}
\newcommand{\problemd}{\textbf{\textit{NK-S1}}}
\newcommand{\probleme}{\textbf{\textit{HIFF}}}
\newcommand{\problemf}{\textbf{\textit{MAXCUT Sparse}}}
\newcommand{\problemg}{\textbf{\textit{MAXCUT Dense}}}
\newcommand{\problemh}{\textbf{\textit{Spin-glass}}}
\newcommand{\problemi}{\textbf{\textit{MAXSAT}}}

\newcommand{\problematext}{$\mathrm{Trap_5^5}$}
\newcommand{\problembtext}{$\mathrm{Trap_5^4}$}
\newcommand{\problemctext}{Bimodal Trap}
\newcommand{\problemdtext}{NK-S1}
\newcommand{\problemetext}{HIFF}
\newcommand{\problemftext}{MAXCUT Sparse}
\newcommand{\problemgtext}{MAXCUT Dense}
\newcommand{\problemhtext}{Spin-glass}
\newcommand{\problemitext}{MAXSAT}

\subsection{GOMEA design choices search results}\label{results_gom}

We found that the best performance of single-population GOMEA is achieved when using \emph{Single-Iteration Hill Climber}, \emph{Forced Improvements}, \emph{Exhaustive Donor Search}, \emph{Filtered Linkage Tree} which is build based on \emph{Normalized Mutual Information}, \emph{FOS sorted in ascending elements size order}, and \emph{Tournament Selection with tournament size 2} applied before linkage model learning, as highlighted in Table~\ref{tab:params}.


With the best hyper-parameter settings, \gomeabest has better performance than the previously published GOMEA version on 7 out of 9 considered problems. 
These results are shown in Table \ref{tab:results_gom}, and scalability plots are presented in Figure \ref{fig:results_bisection}. 
As shown in Table~\ref{tab:results_gom_pvalues}, these differences in performance are statistically significant at $\alpha=0.01$. 
On the MAXSAT and HIFF problems the improvement is approximately of an order of magnitude. Two problems on which performance became worse are Bimodal Concatenated Trap and MAXCUT Dense. We notice that for the Bimodal Trap problem all algorithms of the GOMEA family perform worse than DSMGA-II. We believe that this is due to pairwise mutual information-based dependency learning fails, not optimal mixing itself (i.e., with the right FOS, scalability is excellent). Improving performance for this type of Deceptive Traps is an interesting question for future research. Importantly, the \gomeabest algorithm was able to solve all considered problems with the given constraints while DSMGA-II failed to solve the HIFF, and MAXCUT Dense problems. Therefore, we can say that \gomeabest is an algorithm that can tackle a larger class of non-trivial problems efficiently and it less likely fails to solve a problem. However, we see that on 4 out of 9 problems the performance of DSMGA-II is better (on \problematext, \problemdtext, and \problemetext,~ with statistical significance). 

\begin{figure*}[ht]
    \centering
    \includegraphics[width=\textwidth]{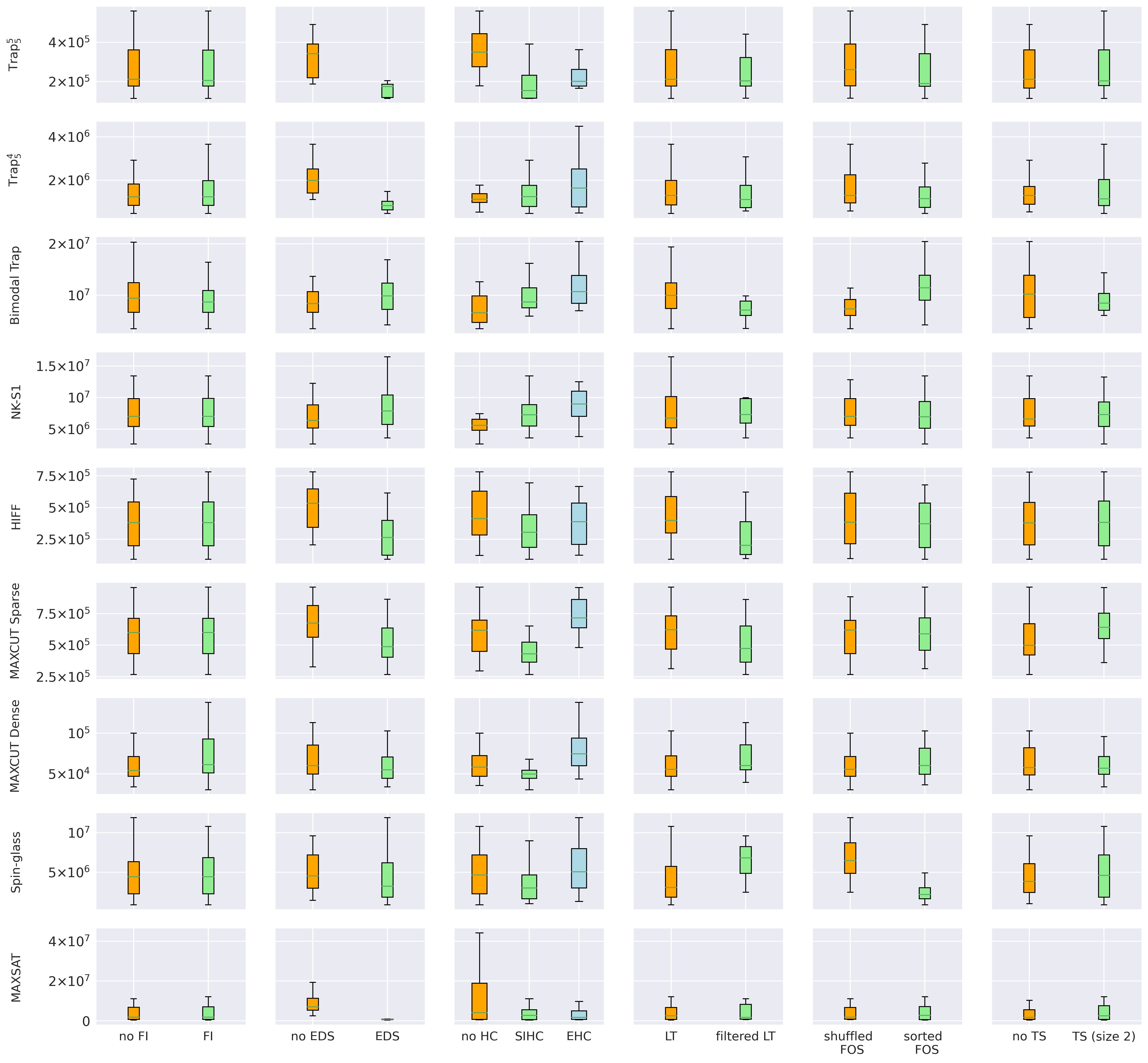}
    \caption{Effects of different design choices in GOMEA. Number of evaluations for each design choice for each problem are aggregated values of all possible GOMEA modifications with this design choice. Population sizes are found with bisection.}
    \label{fig:boxplots}
\end{figure*}

Though the ultimate goal of the conducted hyperparameter search is to find the best performing combination of design choices for GOMEA, it is also interesting to analyze how these choices affect the performance individually. To do so, for each design choice, we study aggregated performance of all algorithms which use this design choice regardless of all other options they use. These results are shown in Figure~\ref{fig:boxplots}. The most impactful design choices are Hill Climber, and Exhaustive Donor Search. Results show that for most problems Exhaustive Donor search is beneficial and substantially improves the performance. 
Algorithms with Single-Iteration Hill Climber on most problems outperform the ones without it, but Exhaustive Hill Climber is, apparently, too greedy and therefore is inferior to both a more simple Hill Climber and no Hill Climber at all. This is in-line with with earlier reported results \cite{bosman2011roles}. Using the Filtered Linkage Tree built with the Normalized Mutual Information measure slightly improves the performance on some problems from the benchmark set, though it worsens the performance on the remaining ones. We see that Forced Improvements, FOS ordering and Tournament Selection do not have a strong effect on the performance. It is noteworthy that the effects of different design choices on the performance of GOMEA on the NK-landscapes are the opposite to their effect on the majority of other problems (e.g., Exhaustive Donor Search, Hill Climber and Filtered LT have worse performance), which suggests that the NK-landscapes problem has some unique properties compared to the other problems in the benchmark set.

\subsection{CGOMEA performance}\label{results_CGOM}
We take the found best performing GOMEA version ($\mathrm{GOMEA^{best}}$) and replace GOM with CGOM. First, we analyze how the performance of CGOM-based GOMEA depends on the threshold parameter $\lambda$. Results for single-population CGOMEA with different $\lambda$ values are provided in Table~\ref{tab:results_CGOM}. We see that $\lambda$ values between 0.6 and 0.9 provide similar performance on most problems though there are some outliers in performance (as on \problemgtext~ problem with $\lambda=0.7$) which are caused by sthe tochastic nature of the bisection procedure. Nevertheless, using the same approach as for selecting the best GOMEA version, we select $\lambda=0.8$ as the value which provides the best average performance. CGOMEA with tuned $\lambda$ value is further referred to as \cgomeabest. We see that with $\lambda=0.5$ performance deteriorates as detecting too many spurious dependencies slows down the mixing procedure. Hence, trying smaller values for $\lambda$ is not necessary.

\begin{figure*}
    \centering
    \includegraphics[height=0.4\textheight]{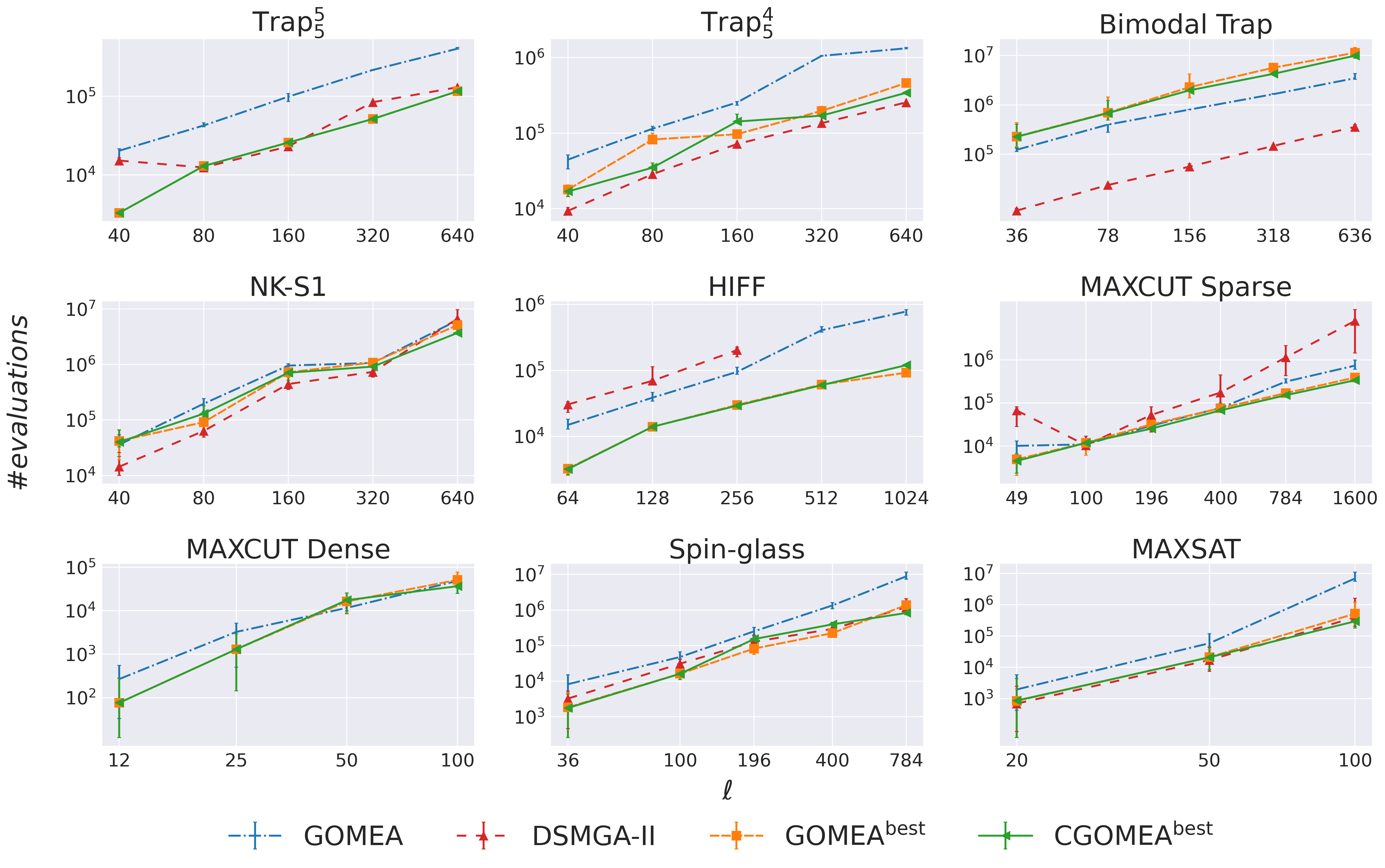}
    \caption{Scalability of single-population EAs in terms of number of function evaluations required to find an optimum. Points show median values of 50 runs. Bars show 3nd and 48th order statistics ($92\%$ confidence interval). If an algorithm fails to find the global optimum of a problem instance in all 50 runs, the corresponding point is not shown. GOMEA refers to the previously published version \cite{bosman2012linkage}.}
    \label{fig:results_bisection}
\end{figure*}

As shown in Table~\ref{tab:results_gom} and in scalability plots in Figure~\ref{fig:results_bisection}, \cgomeabest outperforms \gomeabest on 7 out of 9 considered problems. These differences are statistically significant as shown in Table~\ref{tab:results_gom_pvalues}. On \problematext~ \cgomeabest performs on par with \gomeabest. Only on the HIFF problem CGOM performs slightly worse. Moreover, \cgomeabest performs better (with statistical significance) than DSMGA-II on 5 problems, and there are two problems (HIFF and MAXCUT Dense) which CGOMEA managed to solve but DSMGA-II did not. Importantly, CGOMEA is still able to reliably solve all considered problems. CGOMEA's slightly inferior performance on the HIFF problem can be explained by the complicated structure of HIFF: dependencies exist between all pairs of variables, and CGOM tends to include many variables as dependent ones, leading to less efficient variation as the pool of appropriate donors becomes more limited.

\begin{figure*}
    \centering
    \includegraphics[height=0.4\textheight]{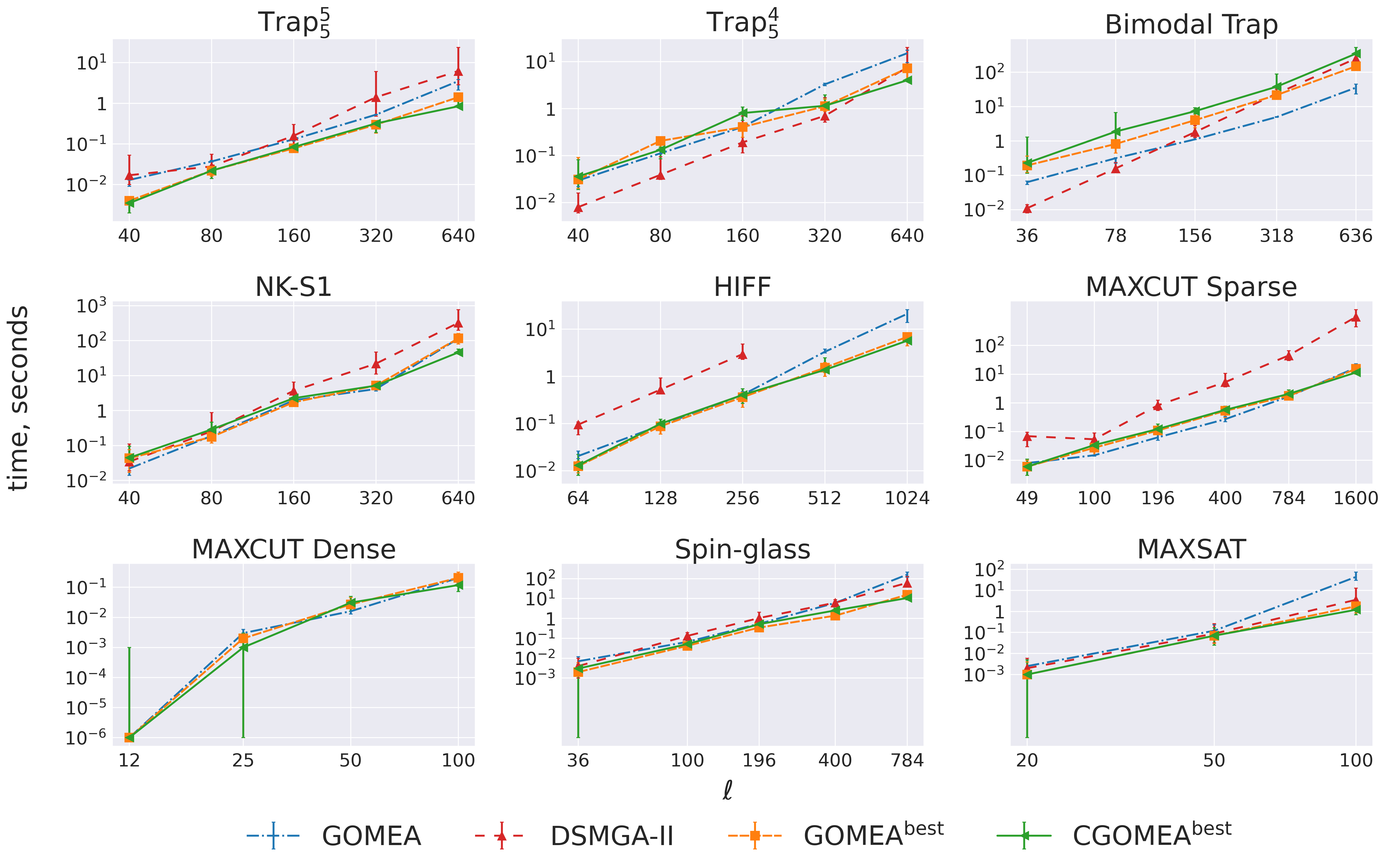}
    \caption{Scalability of single-population EAs in terms of wall clock time required to find an optimum. Points show median values of 50 runs. Bars show 3rd and 48th order statistics ($92\%$ confidence interval). If an algorithm fails to find the global optimum of a problem instance in all 50 runs, the corresponding point is not shown.}
    \label{fig:results_bisection_time}
\end{figure*}

The scalability of single-population algorithms in terms of wall-clock time required to find an optimum  is shown in Figure~\ref{fig:results_bisection_time}. CGOMEA and GOMEA scale similarly on all problems which is better than DSMGA-II, especially on \problematext, \problemctext, \problemdtext, \problemetext, and \problemftext. Only on \problemctext~ CGOMEA is substantially slower than GOMEA though it requires fewer function evaluations. This can be explained by the more careful donor selection as done it CGOMEA. On the NP-hard MAXSAT problem scalability deviates from polynomial as expected, though on the NP-hard MAXCUT Dense problem it cannot be seen for the considered problem sizes.

\begin{table}
\small
\caption{Results of single-population EAs. Best population sizes are found with bisection. Ranking per problem shown through color gradient from green (best, i.e., the fewest median number of function evaluations) to red (worst, i.e., largest median number of function evaluations or problem instance not solved in all 50 runs). All results are divided by $10^5$. Legend: $\mathbf{G} = \mathbf{GOMEA}$; $\mathbf{D\!-\!II} = \mathbf{DSMGA\!-\!II}$; $\mathbf{G^{B}} = \mathbf{GOMEA^{BEST}}$; $\mathbf{CG^{B}} = \mathbf{CGOMEA^{BEST}}$.}
\vspace*{-3mm}
\label{tab:results_gom}
\setlength{\tabcolsep}{1mm}
\begin{center}
\begin{tabular}{c|c|c|c|c|c}
     \textbf{Problem} & $\mathbf{\ell}$ & $\mathbf{G}$ & $\mathbf{D\!-\!II}$ & $\mathbf{G^{B}}$ & $\mathbf{CG^{B}}$ \\
     \hline 
\problema & 640 & \cellcolor[rgb]{1.0, 0.388, 0.278} 4.02 & \cellcolor[rgb]{0.971, 0.834, 0.38} 1.30 & \cellcolor[rgb]{0.565, 0.933, 0.565} 1.16 & \cellcolor[rgb]{0.767, 0.949, 0.476} 1.16 \\
\problemb & 640 & \cellcolor[rgb]{1.0, 0.388, 0.278} 13.23 & \cellcolor[rgb]{0.565, 0.933, 0.565} 2.55 & \cellcolor[rgb]{0.971, 0.834, 0.38} 4.60 & \cellcolor[rgb]{0.767, 0.949, 0.476} 3.41 \\
\problemc & 636 & \cellcolor[rgb]{0.767, 0.949, 0.476} 34.87 & \cellcolor[rgb]{0.565, 0.933, 0.565} 3.60 & \cellcolor[rgb]{1.0, 0.388, 0.278} 114.36 & \cellcolor[rgb]{0.971, 0.834, 0.38} 99.69 \\
\problemd & 640 & \cellcolor[rgb]{0.971, 0.834, 0.38} 62.65 & \cellcolor[rgb]{1.0, 0.388, 0.278} 66.72 & \cellcolor[rgb]{0.767, 0.949, 0.476} 51.36 & \cellcolor[rgb]{0.565, 0.933, 0.565} 37.03 \\
\probleme & 1024 & \cellcolor[rgb]{0.971, 0.834, 0.38} 7.81 & \cellcolor[rgb]{1.0, 0.388, 0.278} N/A & \cellcolor[rgb]{0.565, 0.933, 0.565} 0.92 & \cellcolor[rgb]{0.767, 0.949, 0.476} 1.20 \\
\problemf & 1600 & \cellcolor[rgb]{0.971, 0.834, 0.38} 7.31 & \cellcolor[rgb]{1.0, 0.388, 0.278} 80.03 & \cellcolor[rgb]{0.767, 0.949, 0.476} 3.87 & \cellcolor[rgb]{0.565, 0.933, 0.565} 3.35 \\
\problemg & 100 & \cellcolor[rgb]{0.767, 0.949, 0.476} 0.49 & \cellcolor[rgb]{1.0, 0.388, 0.278} N/A & \cellcolor[rgb]{0.971, 0.834, 0.38} 0.51 & \cellcolor[rgb]{0.565, 0.933, 0.565} 0.37 \\
\problemh & 784 & \cellcolor[rgb]{1.0, 0.388, 0.278} 88.47 & \cellcolor[rgb]{0.767, 0.949, 0.476} 11.92 & \cellcolor[rgb]{0.971, 0.834, 0.38} 13.74 & \cellcolor[rgb]{0.565, 0.933, 0.565} 8.26 \\
\problemi & 100 & \cellcolor[rgb]{1.0, 0.388, 0.278} 69.63 & \cellcolor[rgb]{0.767, 0.949, 0.476} 3.97 & \cellcolor[rgb]{0.971, 0.834, 0.38} 5.20 & \cellcolor[rgb]{0.565, 0.933, 0.565} 2.95 \\
\hline
\end{tabular}
\end{center}
\end{table}

\begin{table}
\small
\caption{Results of single-population CGOMEA with different threshold values $\lambda$. Best population sizes are found with bisection. Ranking per problem shown through color gradient from green (best, i.e., the fewest median number of function evaluations) to red (worst, i.e., largest median number of function evaluations or problem instance not solved in all 50 runs). All results are divided by $10^5$.}
\label{tab:results_CGOM}
\vspace*{-6mm}
\begin{center}
\setlength{\tabcolsep}{0.7mm}
\begin{tabular}{ccccccc}
  \multicolumn{2}{c}{} & \multicolumn{5}{c}{$\boldsymbol{\lambda}$} \\
     \hline 
     \textbf{Problem} & $\ell$ & $\mathbf{0.5}$ & $\mathbf{0.6}$ & $\mathbf{0.7}$ & $\mathbf{0.8}$ & $\mathbf{0.9}$  \\
     \hline 
\problema & 640 & \cellcolor[rgb]{0.565, 0.933, 0.565} 1.16 & \cellcolor[rgb]{0.705, 0.945, 0.499} 1.16 & \cellcolor[rgb]{0.919, 0.959, 0.429} 1.16 & \cellcolor[rgb]{0.978, 0.738, 0.355} 1.16 & \cellcolor[rgb]{1.0, 0.388, 0.278} 1.16 \\
\problemb & 640 & \cellcolor[rgb]{1.0, 0.388, 0.278} 5.89 & \cellcolor[rgb]{0.565, 0.933, 0.565} 3.24 & \cellcolor[rgb]{0.978, 0.738, 0.355} 5.37 & \cellcolor[rgb]{0.705, 0.945, 0.499} 3.41 & \cellcolor[rgb]{0.919, 0.959, 0.429} 3.50 \\
\problemc & 636 & \cellcolor[rgb]{1.0, 0.388, 0.278} 120.73 & \cellcolor[rgb]{0.978, 0.738, 0.355} 106.39 & \cellcolor[rgb]{0.705, 0.945, 0.499} 100.11 & \cellcolor[rgb]{0.565, 0.933, 0.565} 99.69 & \cellcolor[rgb]{0.919, 0.959, 0.429} 104.63 \\
\problemd & 640 & \cellcolor[rgb]{0.705, 0.945, 0.499} 31.96 & \cellcolor[rgb]{0.565, 0.933, 0.565} 28.77 & \cellcolor[rgb]{0.919, 0.959, 0.429} 34.18 & \cellcolor[rgb]{0.978, 0.738, 0.355} 37.03 & \cellcolor[rgb]{1.0, 0.388, 0.278} 46.86 \\
\probleme & 1024 & \cellcolor[rgb]{1.0, 0.388, 0.278} 1.35 & \cellcolor[rgb]{0.705, 0.945, 0.499} 1.10 & \cellcolor[rgb]{0.919, 0.959, 0.429} 1.15 & \cellcolor[rgb]{0.978, 0.738, 0.355} 1.20 & \cellcolor[rgb]{0.565, 0.933, 0.565} 0.91 \\
\problemf & 1600 & \cellcolor[rgb]{0.978, 0.738, 0.355} 3.84 & \cellcolor[rgb]{1.0, 0.388, 0.278} 3.90 & \cellcolor[rgb]{0.565, 0.933, 0.565} 3.27 & \cellcolor[rgb]{0.705, 0.945, 0.499} 3.35 & \cellcolor[rgb]{0.919, 0.959, 0.429} 3.46 \\
\problemg & 100 & \cellcolor[rgb]{0.978, 0.738, 0.355} 0.62 & \cellcolor[rgb]{0.705, 0.945, 0.499} 0.53 & \cellcolor[rgb]{1.0, 0.388, 0.278} 0.62 & \cellcolor[rgb]{0.565, 0.933, 0.565} 0.37 & \cellcolor[rgb]{0.919, 0.959, 0.429} 0.58 \\
\problemh & 784 & \cellcolor[rgb]{1.0, 0.388, 0.278} 16.49 & \cellcolor[rgb]{0.919, 0.959, 0.429} 11.98 & \cellcolor[rgb]{0.978, 0.738, 0.355} 12.45 & \cellcolor[rgb]{0.565, 0.933, 0.565} 8.26 & \cellcolor[rgb]{0.705, 0.945, 0.499} 10.78 \\
\problemi & 100 & \cellcolor[rgb]{0.978, 0.738, 0.355} 4.08 & \cellcolor[rgb]{1.0, 0.388, 0.278} 4.89 & \cellcolor[rgb]{0.919, 0.959, 0.429} 3.44 & \cellcolor[rgb]{0.565, 0.933, 0.565} 2.95 & \cellcolor[rgb]{0.705, 0.945, 0.499} 3.25 \\
\hline
\end{tabular}
\end{center}
\end{table}

\subsection{Parameterless EAs}
\begin{figure*}
    \centering
    \includegraphics[height=0.4\textheight]{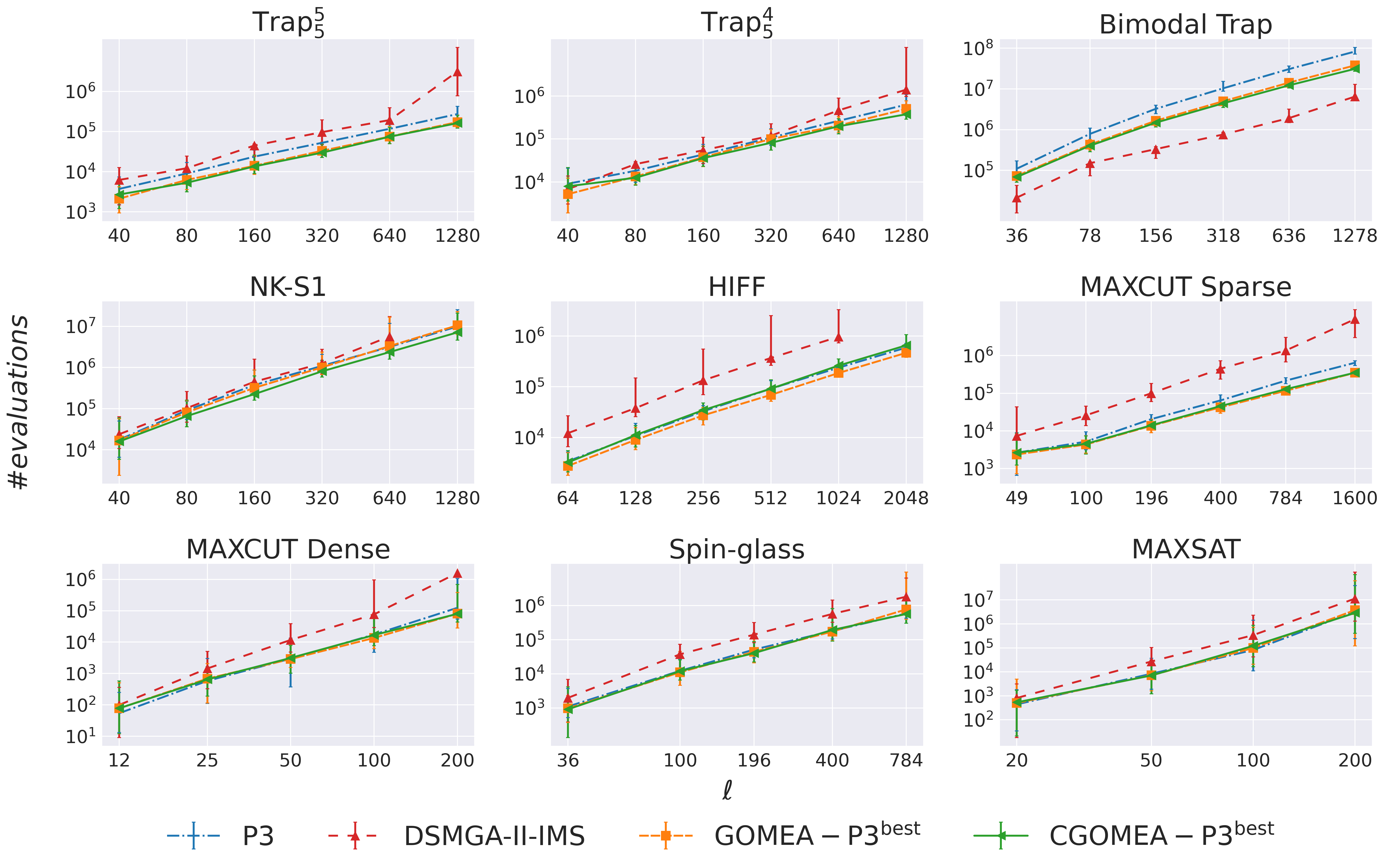}
    \caption{Scalability of parameterless EAs in terms of number of function evaluations required to find an optimum. Points show median values of 50 runs (48 runs for MAXSAT problem). Bars show 3nd and 48th (46th for MAXSAT problem) order statistics ($92\%$ confidence interval). If an algorithm fails to find the global optimum of a problem instance in all 50 runs (48 runs for MAXSAT problem), the corresponding point is not shown.}
    \label{fig:results_popfree}
\end{figure*}

Results of experiments with parameterless EAs are presented in Table~\ref{tab:results_popfree} and in scalability plots in Figure~\ref{fig:results_popfree}. We found that the best performance of a parameterless GOMEA is achieved when GOMEA uses \emph{Single-Iteration Hill Climber}, \emph{Exhaustive Donor Search}, \emph{Filtered Linkage Tree} which is build based on \emph{Normalized Mutual Information}, \emph{randomly shuffled FOS}, and \emph{P3} scheme, as highlighted in Table~\ref{tab:hyperparams_popfree}. The obtained parameterless GOMEA version is further referred to as \gomeabestp. Note, that when P3 and P3-MI schemes do not use tournament selection, it makes them much more time efficient, as population statistics needed for Linkage Learning can be efficiently updated instead of re-calculated from scratch \cite{goldman2015fast}. Noteworthy, crucial design choices, such as Hill Climber, Exhaustive Donor Search, and Linkage Tree type and information measure are the same in \gomeabest and \gomeabestp. Less important design choices (Forced Improvements, FOS ordering) differ, which is most likely due to the results stochastic nature. The \gomeabestp version, but with GOM replaced by CGOM ($\lambda=0.8$ corresponding to the best value found in Section~\ref{hyperparamsGOM}) is further referred to as \cgomeabestp.

First, we see that DSMGA-II with IMS scheme was capable to solve only three problems in the experimental setup due to its issues with time efficiency. 
\cgomeabestp performs better than \gomeabestp on 5 problems out of 9  (on 4 of them with statistical significance). The most substantial differences are on \problemctext, \problemdtext ,and \problembtext. Similar to results for single-population algorithms, CGOMEA performs worse than GOMEA on the HIFF problem, and differences between CGOMEA and GOMEA on \problematext,~ \problembtext,~ and \problemftext are subtle. Compared to the P3 algorithm \cite{goldman2015fast} \cgomeabestp performs better on all problems except \problemetext~(on 6 problems with statistical significance). \gomeabestp performs better than P3 on 6 problems (with statistical significance). 

We note that only DSMGA-II with IMS was capable of solving all 50 instances of the MAXSAT problem of size 200 within the given time limit. \cgomeabestp and P3 solved 49 problem instances, while \gomeabestp solved 48. As problem instances significantly vary in complexity, we show the results for the 48 instances that were solved by all algorithms in order to provide a fair comparison.

\begin{figure*}
    \centering
    \includegraphics[height=0.4\textheight]{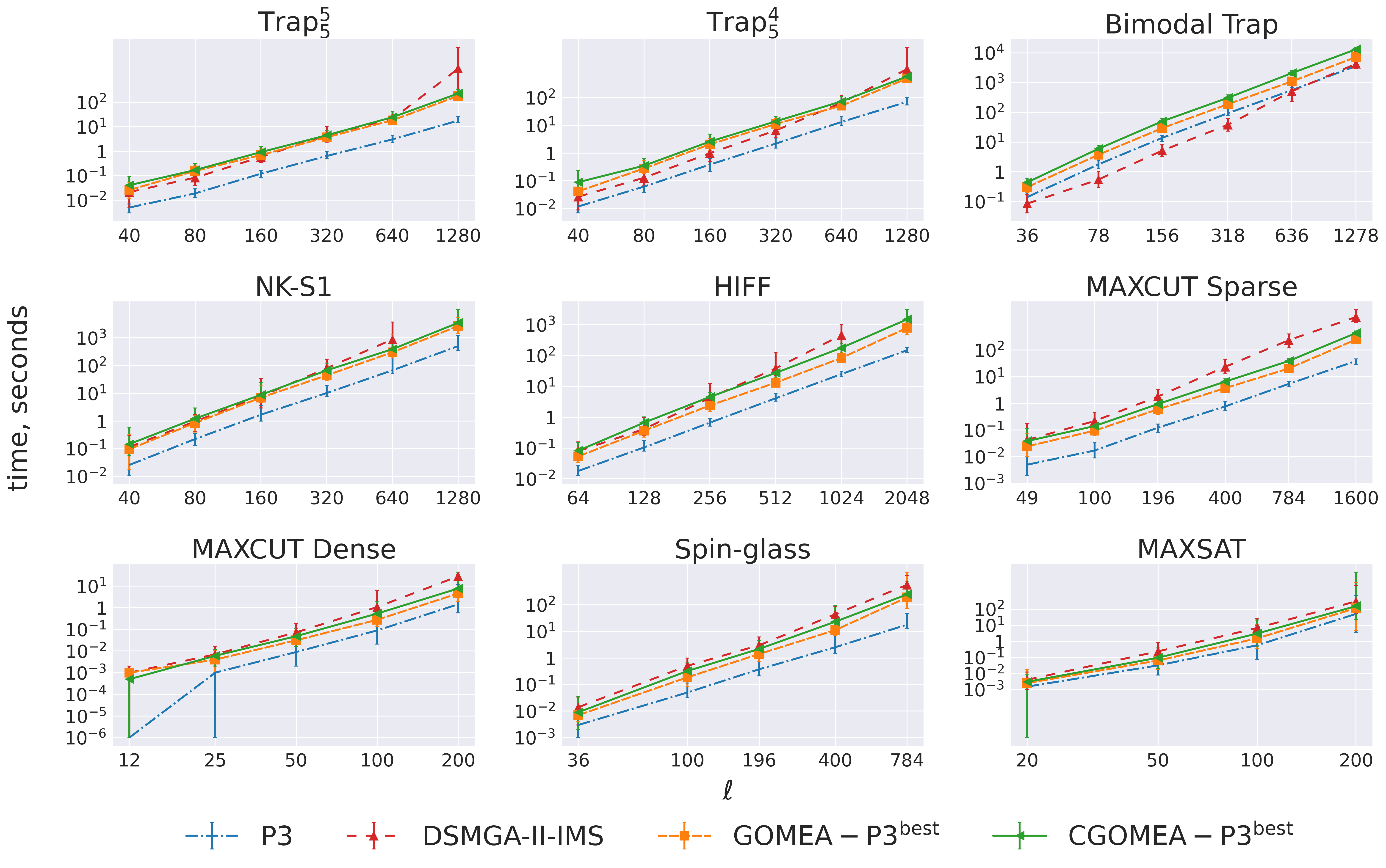}
     \caption{Scalability of parameterless EAs in terms of wall clock time required to find an optimum. Points show median values of 50 runs (48 runs for MAXSAT problem). Bars show 3nd and 48th (46th for MAXSAT problem) order statistics ($92\%$ confidence interval). If an algorithm fails to find the global optimum of a problem instance in all 50 runs (48 runs for MAXSAT problem), the corresponding point is not shown.}
    \label{fig:results_popfree_time}
\end{figure*}

As shown in Figure~\ref{fig:results_popfree_time}, P3 versions of GOMEA and CGOMEA scale similarly to P3, though they are slower. Scalability in terms of required time to find an optimum is almost identical for \cgomeabestp and \gomeabestp. Both \cgomeabestp and \gomeabestp scale better than DSMGA-II IMS on most problems.

\begin{table}
\small
\caption{Results of parameterless EAs. Ranking per problem shown through color gradient from green (best, i.e., fewest median number of function evaluations) to red (worst, i.e., largest median number of function evaluations or problem instance not solved in all 50 runs). For MAXSAT results are shown for 48 instances that were solved by all algorithms. All results are divided by $10^5$. Legend: $\mathbf{P3} = \mathbf{P3}$; $\mathbf{D\!-\!II} = \mathbf{DSMGA\!-\!II}$; $\mathbf{GP3^{B}} = \mathbf{GOMEA\!-\!P3^{BEST}}$; $\mathbf{CGP3^{B}} = \mathbf{CGOMEA\!-\!P3^{BEST}}$.}
\label{tab:results_popfree}
\vspace*{-3mm}
\begin{center}
\setlength{\tabcolsep}{1mm}
\begin{tabular}{c|c|c|c|c|c}
     \textbf{Problem} & $\mathbf{\ell}$ & $\mathbf{P3}$ & $\mathbf{D\!-\!II}$ & $\mathbf{GP3^{B}}$ & $\mathbf{CGP3^{B}}$ \\
     \hline 
\problema & 1280 & \cellcolor[rgb]{0.971, 0.834, 0.38} 2.72 & \cellcolor[rgb]{1.0, 0.388, 0.278} 31.10 & \cellcolor[rgb]{0.767, 0.949, 0.476} 1.71 & \cellcolor[rgb]{0.565, 0.933, 0.565} 1.61 \\
\problemb & 1280 & \cellcolor[rgb]{0.971, 0.834, 0.38} 6.29 & \cellcolor[rgb]{1.0, 0.388, 0.278} 14.08 & \cellcolor[rgb]{0.767, 0.949, 0.476} 5.01 & \cellcolor[rgb]{0.565, 0.933, 0.565} 3.76 \\
\problemc & 1278 & \cellcolor[rgb]{1.0, 0.388, 0.278} 829.32 & \cellcolor[rgb]{0.565, 0.933, 0.565} 65.03 & \cellcolor[rgb]{0.971, 0.834, 0.38} 379.01 & \cellcolor[rgb]{0.767, 0.949, 0.476} 311.83 \\
\problemd & 1280 & \cellcolor[rgb]{0.767, 0.949, 0.476} 99.10 & \cellcolor[rgb]{1.0, 0.388, 0.278} N/A & \cellcolor[rgb]{0.971, 0.834, 0.38} 105.58 & \cellcolor[rgb]{0.565, 0.933, 0.565} 71.21 \\
\probleme & 2048 & \cellcolor[rgb]{0.767, 0.949, 0.476} 5.92 & \cellcolor[rgb]{1.0, 0.388, 0.278} N/A & \cellcolor[rgb]{0.565, 0.933, 0.565} 4.67 & \cellcolor[rgb]{0.971, 0.834, 0.38} 6.55 \\
\problemf & 1600 & \cellcolor[rgb]{0.971, 0.834, 0.38} 6.47 & \cellcolor[rgb]{1.0, 0.388, 0.278} 93.12 & \cellcolor[rgb]{0.565, 0.933, 0.565} 3.50 & \cellcolor[rgb]{0.767, 0.949, 0.476} 3.52 \\
\problemg & 200 & \cellcolor[rgb]{0.971, 0.834, 0.38} 1.24 & \cellcolor[rgb]{1.0, 0.388, 0.278} 15.97 & \cellcolor[rgb]{0.767, 0.949, 0.476} 0.80 & \cellcolor[rgb]{0.565, 0.933, 0.565} 0.80 \\
\problemh & 784 & \cellcolor[rgb]{0.767, 0.949, 0.476} 5.84 & \cellcolor[rgb]{1.0, 0.388, 0.278} 18.15 & \cellcolor[rgb]{0.971, 0.834, 0.38} 7.69 & \cellcolor[rgb]{0.565, 0.933, 0.565} 5.65 \\
\problemi & 200 & \cellcolor[rgb]{0.767, 0.949, 0.476} 32.43 & \cellcolor[rgb]{1.0, 0.388, 0.278} 113.84 & \cellcolor[rgb]{0.971, 0.834, 0.38} 37.57 & \cellcolor[rgb]{0.565, 0.933, 0.565} 28.74 \\
     \hline
\end{tabular}
\end{center}
\end{table}

\section{Discussion}\label{sec:discussion}
We implemented the conditional GOM operator using traditional, entropy-based similarity measures to predict dependencies between variables. 
Especially in early generations, this approach to detecting dependencies can be inaccurate, determining dependencies between variables which are actually independent, and missing some truly existing ones. Potentially, a more accurate approach to learning dependencies can further improve CGOM performance. 
Moreover, it can be interesting to apply CGOM operator in a Gray-Box Optimization (GBO) scenario when the true dependencies are known. Then, as was done for RV-GOMEA, a Bayesian network could be used rather than the conditional variant of the LT. The latter is advantageous when learning linkage in a BBO setting, but not as accurate and potentially more complex compared to a direct and concise modelling of conditional dependencies. This analogy of the original concept of CGOM is to be studied in future research. However, in that case, it should be compared to different forms of EAs such as \cite{chicano2017optimizing} which were designed specifically for GBO. 


The considered model-based EAs relied on entropy-based information measures to learn dependencies between variables. We notice that the performance on \problemctext~ can be potentially improved if alternative linkage learning methods are used, such as fitness-based ones as it is known that alternative methods that use comparisons can find the right structure \cite{przewozniczek2020empirical, telo}. In general, the current state-of-the-art results are achieved by entropy-based linkage learning techniques, though, replacing them or combining with other methods is a promising question for future research.

In this article, to determine the best design choices (hyperparameters) for GOMEA and CGOMEA, we assessed performance on a standard benchmark set, and ranked algorithms based on average performance.
Though this benchmark set includes well-known combinatorial optimization problems, problems arising in practical tasks may have properties (such as fitness landscape and dependencies structure) which are very different from all common benchmark functions. 
Though practitioners are interested in having the best performing algorithm for their specific task, we do not have a priori knowledge of those tasks properties. Defining a good and comprehensive benchmark set is an open problem and active field of research \cite{van2020towards}. We hypothesize,  however that the state-of-the-art benchmark problems in the field of EAs for binary optimization that we used is a decent compromise, in that we expect that obtaining good average performance on these problems is a good predictor of performance on many a priori unknown tasks. 
Moreover, in a BBO scenario matching a real-world problem with a problem from a benchmark set is a hard, if even solvable, task itself. Therefore, we did not try to specify the best possible GOMEA and CGOMEA versions for each benchmark problem, but keep the focus on the best average performance.

\section{Conclusion}\label{sec:conclusion}
In this article we have continued the research line on the GOMEA family of algorithms with important innovations and comparisons of various ideas that have been proposed separately in the last decade since the introduction of GOMEA. First, we did an extensive hyperparameter search and obtained a version of GOMEA which showed significantly better performance than ever published before for GOMEA. 
Next, we introduced a new variation operator called Conditional Gene-pool Optimal Mixing (CGOM) which utilizes conditional dependencies of linkage model subsets on other variables to generate offspring solutions. 
GOMEA with CGOM (CGOMEA) outperformed GOMEA and DSMGA-II on most of the 9 considered diverse and non-trivial benchmark problems in a single-population EA experimental setup where we assess scalability of the algorithms of required resources to obtain the optimum. 
Finally, we searched for the best performing version of GOMEA integrated with various population size-free schemes. 
We found that CGOMEA with P3 scheme is a robust scalable algorithm which outperforms the competitors in terms of number of function evaluations required to find the global optimum on almost all problems setting a new state-of-the-art performance for most of the benchmark problems and a new GOMEA variant that can serve as a new baseline in model-based evolutionary algorithms for binary search spaces for the next decade.



\definecolor{first}{RGB}{0,150,0}
\definecolor{second}{RGB}{0,150,0}
\definecolor{third}{RGB}{0,150,0}

\colorlet{third}{BurntOrange!50}
\colorlet{fourth}{BrickRed!50}
\colorlet{significant}{Blue!30}

\newcolumntype{Y}{>{\centering\arraybackslash}X}



\clearpage
\begin{table}[ht]

\small
\caption{Statistical significance testing of performance difference in pairs of single-population EAs. Reported $p-$values are obtained by pairwise Mann-Whitney $U$-test. Statistically significant results at $\alpha=0.01$ are highlighted. Cells corresponding to cases when an algorithm solves a problem instance and its competitor does not, are not highlighted. Legend: $\mathbf{G} = \mathbf{GOMEA}$; $\mathbf{D\!-\!II} = \mathbf{DSMGA\!-\!II}$; $\mathbf{G^{B}} = \mathbf{GOMEA\!-\!^{BEST}}$; $\mathbf{CG^{B}} = \mathbf{CGOMEA\!-\!^{BEST}}$.}

\label{tab:results_gom_pvalues}
\setlength{\tabcolsep}{1mm}
\begin{tabularx}{0.5\textwidth}{c|c|X|X|X|X}
     \textbf{Problem} & $\ell$ & $\mathbf{G^{B}}$ is better than $\mathbf{G}$ & $\mathbf{G^{B}}$ is better than $\mathbf{D\!-\!II}$ & $\mathbf{CG^{B}}$ is better than $\mathbf{G^{B}}$ & $\mathbf{CG^{B}}$ is better than $\mathbf{D\!-\!II}$ \\
     \hline 
\problema & 640 & \cellcolor{significant} 0.000 & \cellcolor{significant} 0.000 & 0.503 & \cellcolor{significant} 0.000 \\
\problemb & 640 & \cellcolor{significant} 0.000 & 1.000 & \cellcolor{significant} 0.000 & 1.000 \\
\problemc & 636 & 1.000 & 1.000 & \cellcolor{significant} 0.000 & 1.000 \\
\problemd & 640 & \cellcolor{significant} 0.000 & \cellcolor{significant} 0.000 & \cellcolor{significant} 0.000 & \cellcolor{significant} 0.000 \\
\probleme & 1024 & \cellcolor{significant} 0.000 & N/A & 1.000 & N/A \\
\problemf & 1600 & \cellcolor{significant} 0.000 & \cellcolor{significant} 0.000 & \cellcolor{significant} 0.000 & \cellcolor{significant} 0.000 \\
\problemg & 100 & 0.740 & N/A & \cellcolor{significant} 0.000 & N/A \\
\problemh & 784 & \cellcolor{significant} 0.000 & 1.000 & \cellcolor{significant} 0.000 & \cellcolor{significant} 0.000 \\
\problemi & 100 & \cellcolor{significant} 0.000 & 0.960 & \cellcolor{significant} 0.000 & \cellcolor{significant} 0.000 \\
\hline
\end{tabularx}
\vspace{2cm}
\small
\label{tab:popfree_pvalues}
\caption{Statistical significance testing of performance difference in pairs of the best performing parameterless EAs. Reported $p-$values are obtained by pairwise Mann-Whitney $U$-test. Statistically significant results at $\alpha=0.01$ are highlighted. Legend: $\mathbf{P3} = \mathbf{P3}$; $\mathbf{D\!-\!II} = \mathbf{DSMGA\!-\!II}$; $\mathbf{GP3^{B}} = \mathbf{GOMEA\!-\!P3^{BEST}}$; $\mathbf{CGP3^{B}} = \mathbf{CGOMEA\!-\!P3^{BEST}}$.}
\setlength{\tabcolsep}{1mm}
\begin{tabularx}{0.5\textwidth}{c|c|X|X|X}
     \textbf{Problem} & $\ell$ & $\mathbf{CGP3^{B}}$ is better than $\mathbf{GP3^{B}}$
     & $\mathbf{CGP3^{B}}$ is better than $\mathbf{P3}$
     & $\mathbf{GP3^{B}}$ is better than $\mathbf{P3}$
     \\ \hline
     \problema & 1280 & 0.213 & \cellcolor{significant} 0.000 & \cellcolor{significant} 0.000 \\
\problemb & 1280 & \cellcolor{significant} 0.000 & \cellcolor{significant} 0.000 & \cellcolor{significant} 0.000 \\
\problemc & 1278 & \cellcolor{significant} 0.000 & \cellcolor{significant} 0.000 & \cellcolor{significant} 0.000 \\
\problemd & 1280 & \cellcolor{significant} 0.000 & \cellcolor{significant} 0.000 & 0.662 \\
\probleme & 2048 & 1.000 & 0.997 & \cellcolor{significant} 0.000 \\
\problemf & 1600 & 0.537 & \cellcolor{significant} 0.000 & \cellcolor{significant} 0.000 \\
\problemg & 200 & 0.769 & \cellcolor{significant} 0.003 & \cellcolor{significant} 0.001 \\
\problemh & 784 & \cellcolor{significant} 0.007 & 0.227 & 0.985 \\
\problemi & 200 & 0.608 & 0.501 & 0.507 \\
     \hline 

\end{tabularx}
\end{table}

\clearpage
\bibliographystyle{plain}
\bibliography{bibliography}
\end{document}